\documentclass[11pt]{article}

\usepackage[preprint]{acl}

\usepackage{times}
\usepackage{latexsym}

\usepackage[T1]{fontenc}

\usepackage[utf8]{inputenc}

\usepackage{microtype}

\usepackage{inconsolata}

\usepackage{graphicx}
\usepackage{algorithm}
\usepackage{algpseudocode}
\usepackage{amsmath,amssymb}   
\usepackage{graphicx}           
\usepackage{xcolor}             
\usepackage{booktabs}       
\usepackage{amsmath}
\usepackage{cases}
\usepackage{bm}
\usepackage{amsmath}
\usepackage{multirow}
\usepackage{cuted}

\usepackage{graphicx}      
\usepackage{booktabs}      
\usepackage{multirow}      
\usepackage{subcaption}    
\usepackage[table]{xcolor} 
\usepackage{tcolorbox}     
\usepackage{fancyvrb}
\usepackage[dvipsnames]{xcolor}

\usepackage{hyperref}
\usepackage{url}
\usepackage{tikz}
\usepackage{tabularx}
\usepackage{booktabs}       
\usepackage{multirow}
\usepackage{seqsplit}
\usepackage{makecell}
\usepackage{listings}
\usepackage{enumitem}
\usepackage{xcolor}
\usepackage{alltt}
\usepackage{fancyvrb}
\usepackage[utf8]{inputenc}
\newcolumntype{Y}{>{\raggedright\arraybackslash}X}
\DeclareMathOperator{\Valid}{\operatorname{Pass}}

\definecolor{gaincolor}{HTML}{2A6099} 
\definecolor{losscolor}{HTML}{B32428} 

\title{TAROT: Test-driven and Capability-adaptive Curriculum Reinforcement Fine-tuning for Code Generation with Large Language Models}

\author{Chansung Park$^1$\thanks{Equal contributors: Juyong Jiang, Chansung Park, and Fan Wang. Listing order is random.}\quad 
Juyong Jiang$^2$\,$^3$\footnotemark[1]\quad
Fan Wang$^2$\footnotemark[1]\quad
Sayak Paul$^4$\\
\textbf{Jiasi Shen}$^3$\thanks{Corresponding authors.}\quad  
\textbf{Jing Tang}$^2$\,$^3$\footnotemark[2]\quad
\textbf{Jianguo Li}$^5$\footnotemark[2]\quad \\
$^1$Electronics and Telecommunications Research Institute \\
$^2$The Hong Kong University of Science and Technology (Guangzhou) \\
$^3$The Hong Kong University of Science and Technology 
$^4$Hugging Face $^5$Ant Group \\
\texttt{\{deep.diver.csp,csjuyongjiang,csfanwang,spsayakpaul\}@gmail.com}\\
\texttt{sjs@cse.ust.hk},
\texttt{jingtang@ust.hk},
\texttt{lijg.zero@antgroup.com}
}

\begin{document}
\maketitle
\begin{abstract}
Large Language Models (LLMs) are changing the coding paradigm, known as vibe coding, yet synthesizing algorithmically sophisticated and robust code still remains a critical challenge. Incentivizing the deep reasoning capabilities of LLMs is essential to overcoming this hurdle. Reinforcement Fine-Tuning (RFT) has emerged as a promising strategy to address this need. However, most existing approaches overlook the heterogeneous difficulty and granularity inherent in test cases, leading to an imbalanced distribution of reward signals and consequently biased gradient updates during training. To address this, we propose \textbf{T}est-driven and c\textbf{A}pability-adaptive cu\textbf{R}riculum reinf\textbf{O}rcement fine-\textbf{T}uning \textbf{(TAROT)}. TAROT systematically constructs, for each problem, a four-tier test suite (basic, intermediate, complex, edge), providing a controlled difficulty landscape for curriculum design and evaluation. Crucially, TAROT decouples curriculum progression from raw reward scores, enabling capability-conditioned evaluation and principled selection from a portfolio of curriculum policies rather than incidental test-case difficulty composition. This design fosters stable optimization and more efficient competency acquisition. Extensive experimental results reveal that the optimal curriculum for RFT in code generation is closely tied to a model's inherent capability, with less capable models achieving greater gains with an easy-to-hard progression, whereas more competent models excel under a hard-first curriculum. TAROT provides a reproducible method that adaptively tailors curriculum design to a model's capability, thereby consistently improving the functional correctness and robustness of the generated code. All code and data are released to foster reproducibility and advance community research at \texttt{\url{https://github.com/deep-diver/TAROT}}.
\end{abstract}

\section{Introduction}

Large Language Models (LLMs) are driving significant changes in software engineering, with automated code generation emerging as a pivotal application~\citep{10.1145/3597503.3639219, jiang2024surveylargelanguagemodels, xu2022survey,park2025llamaduo,wang2024kasa}. Foundational models exhibit a strong capacity to translate natural language specifications into functional code, promising significant enhancements in developer productivity~\citep{10.1145/3661145,li2025osvbench}. Nevertheless, generating solutions that are not only correct but also algorithmically sophisticated and robust remains an open challenge~\citep{zhuo2025bigcodebenchbenchmarkingcodegeneration,zhuo2025bigcodearena}. Advancing this frontier calls for enhancing the reasoning and problem-solving capacities of these models.

Curriculum Learning (CL), which structures training data by difficulty~\citep{bengio_curriculum_2009}, offers a promising pathway for improving these capabilities and training efficiency~\citep{zhang2025learning, nair-etal-2024-curriculum}. In the context of code generation, however, existing CL applications predominantly operate at the inter-problem level, sequencing tasks by coarse difficulty metrics~\citep{nair-etal-2024-curriculum, 11025707}. This curriculum neglects the nuanced, intra-problem difficulty gradient inherent in software verification. In practice, developers often employ incremental methodologies like Test-Driven Development (TDD)~\citep{beck2003test}, refining a solution against progressively more challenging test cases to ensure its robustness. Yet, this natural axis for curriculum design remains largely untapped in LLM training. 
Furthermore, reliance on problem-level sequencing often leads to flat reward landscapes when integrated with Reinforcement Fine-tuning (RFT), dampening the learning signal. This oversight of heterogeneous test-case difficulty results in imbalanced reward signals and consequently biased gradient updates during training, hindering the model's ability to acquire robust, sophisticated reasoning skills.

Recent work has begun to move toward more dynamic curriculum learning paradigms for LLMs, where task difficulty is progressively increased during training~\citep{xu2024wizardlm, cheng2025evocurrselfevolvingcurriculumbehavior}. However, these methods predominantly define difficulty based on the intrinsic properties of the data or task structure. For instance, curricula are often structured using automated metrics like cyclomatic complexity~\citep{nair-etal-2024-curriculum}, or by decomposing a problem into a fixed sequence of simpler subtasks~\citep{dou-etal-2024-stepcoder}. This prevailing focus on the data, rather than the learner, overlooks the crucial variable of the model's own evolving and multi-faceted capability. A curriculum tailored to an early-stage model may cause learning stagnation for a more advanced one, while a curriculum designed for experts can overwhelm a less-capable model and hinder its convergence. A more holistic and effective learning approach, therefore, should consider not just the intrinsic properties of the data, but also the evolving capabilities of the learner itself, leading to a capability-adaptive framework.

To address these limitations, we introduce \textbf{TAROT}, a novel framework for \textbf{T}est-driven and c\textbf{A}pability-adaptive cu\textbf{R}riculum reinf\textbf{O}rcement fine-\textbf{T}uning. TAROT decouples curriculum progression from raw rewards, enabling capability-conditioned evaluation and principled selection from a portfolio of curriculum policies rather than relying on the incidental composition of test-case difficulty. This design fosters stable optimization and efficient competency acquisition. 

Our proposed framework's core novelty is twofold. First, TAROT formalizes an intra-problem difficulty gradient through a test-driven curriculum. We construct the TAROT dataset, where each programming problem is augmented with a tiered test suite comprising basic, intermediate, complex, and edge cases. This structure defines difficulty as a spectrum of functional correctness and allows curriculum progression through differential emphasis on test tiers during training. This engineered gradient provides a structured and informative signal to mitigate the flat-reward issue observed in reinforcement learning (RL) for code generation. Second, TAROT enables a capability-adaptive curriculum design. Leveraging the tiered test structure, TAROT instantiates a portfolio of curriculum policies that differ in tier allocation, sequencing, and reward weighting. This setup supports capability-conditioned evaluation and principled curriculum selection for models with varying effective capabilities, influenced by model scale and specialization. 

Extensive evaluations on state-of-the-art LLMs across coding benchmarks demonstrate that TAROT consistently outperforms the baselines, efficiently enhancing both functional correctness and robustness of the synthesized code. Our empirical investigation further reveals that the optimal curriculum is capability-dependent for RFT in code generation. Specifically, less-capable models benefit from a basic-to-complex progression, whereas more-capable models learn more effectively from complex tiers. 

Our main contributions are as follows:
\begin{itemize}
    \item We propose a capability-adaptive curriculum framework, TAROT, which addresses the imbalanced reward issue in RFT for code generation. It integrates an intra-problem, test-driven dataset featuring four-tiered test suites with a multi-dimensional curriculum portfolio. By providing a granular difficulty landscape for capability-conditioned curriculum design and evaluation, TAROT enables stable and efficient optimization.
    \item We conduct extensive experiments on state-of-the-art LLMs across well-known coding benchmarks. Experimental results show that TAROT consistently improves model performance and training efficiency compared to strong baselines.
    \item Through empirical analysis, we uncover the capability-dependent optimal curriculum. These findings provide guidance for future work on test suite design and automated curriculum selection.
\end{itemize}

\section{Related Work}
\subsection{Curriculum Learning for Code}
Curriculum Learning (CL) is a training strategy inspired by human cognition that presents data to a model in a structured order, typically from simple to complex examples~\citep{bengio_curriculum_2009,wang2021survey}. This method has been shown to accelerate convergence and improve generalization by guiding the optimization process toward better solutions~\citep{ryu2025curricullm,xi2024training}. In the context of LLMs, curricula have been implemented in various ways, such as using a teacher model to progressively generate more complex instructions, as exemplified by the Evol-Instruct method~\citep{xu2024wizardlm}, or by fine-tuning on a small set of meticulously curated, high-quality examples as demonstrated by LIMA~\citep{10.5555/3666122.3668522}.
For code generation, where task complexity varies, CL is a promising but challenging area. While many code datasets rely on manual difficulty labels, recent research has focused on more systematic approaches. A notable example is the use of automatic difficulty metrics, combining measures like cyclomatic complexity and Halstead difficulty, to sort problems into a multi-stage curriculum~\citep{nair-etal-2024-curriculum}. Training with this structured approach yielded significant gains, demonstrating the value of CL in the code domain. Other methods, like StepCoder, create an implicit curriculum by breaking a complex problem into a sequence of simpler code-completion subtasks~\citep{dou-etal-2024-stepcoder}. These efforts show a clear trend towards leveraging curricula to organize the training process for code generation. Unlike these methods, TAROT introduces a tiered test suite to construct a capability-adaptive curriculum for RL.

\subsection{Reinforcement Learning for Code LLMs}

Reinforcement Learning (RL) is widely used for aligning LLMs with desired behaviors, including RLHF~\citep{10.5555/3600270.3602281}, DPO~\citep{10.5555/3666122.3668460}, PPO~\citep{schulman2017proximal}, GRPO~\citep{shao2024deepseekmathpushinglimitsmathematical}, and GSPO~\citep{zheng2025groupsequencepolicyoptimization}. For code generation, RL typically optimizes functional correctness using unit test outcomes as rewards. Despite its effectiveness, this paradigm suffers from two key limitations: reward sparsity and reward flatness. Reward sparsity stems from the lack of informative feedback when a model fails a task entirely, whereas reward flatness occurs when the signal fails to differentiate among levels of problem difficulty~\citep{parashar2025curriculum}. These can lead to imbalanced gradients and suboptimal learning dynamics. Recent work addresses these issues from different angles. Process Reward Models alleviate reward sparsity by providing dense, line-level feedback, guiding the model even when the final code is incorrect ~\citep{dai2025processsupervisionguidedpolicyoptimization}. 
Another line of works combine RL with curriculum design to address these issues. For instance, StepCoder decomposes long tasks into a curriculum of easier subtasks and trains the model step-by-step~\citep{dou2024stepcoder}, while Self-Evolving Curriculum adaptively adjusts curriculum selection, formulated as a multi-armed bandit problem, according to the model’s evolving capability~\citep{chen2025selfevolvingcurriculumllmreasoning}.

Our TAROT framework addresses reward flatness by making the reward signal curriculum-aware. Instead of treating all successes equally, we modulate the reward based on the difficulty of the solved test tier. By integrating this tiered scheme directly into a stable policy optimization algorithm, TAROT provides a more nuanced learning gradient that encourages the model to master harder problems. Integrating a structured curriculum directly into the RL reward mechanism is a novel contribution that complements other recent innovations in the field.

\section{Methodology}
In this section, we elaborate on the proposed TAROT framework, which enhances the code generation capability by training LLMs on test cases of varying difficulty levels, where curriculum progression and reward weights are adaptively conditioned on the model’s baseline capability.

\subsection{Dataset}

\begin{figure*}[t]
\centering
\includegraphics[width=\textwidth]{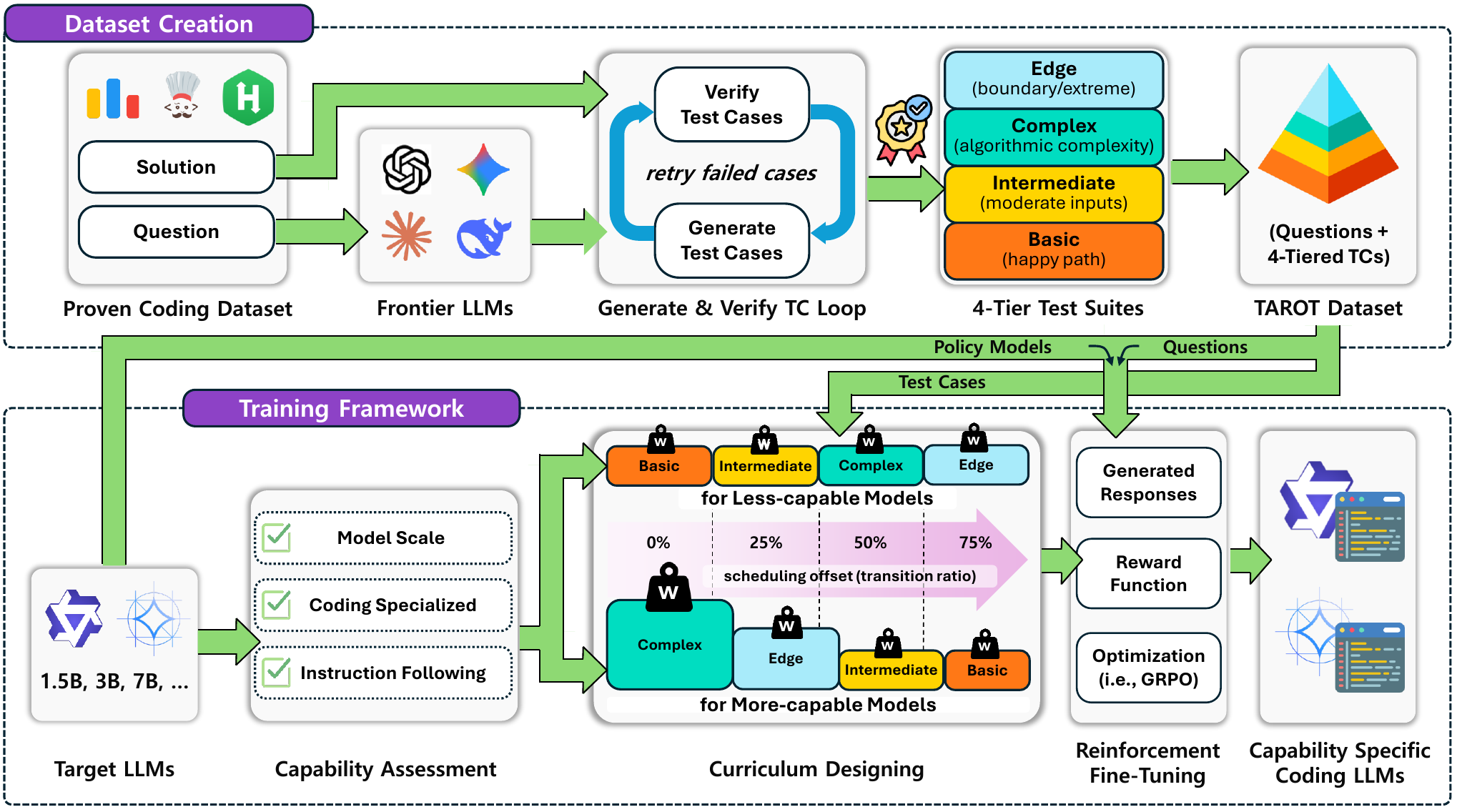}
\caption{\textbf{Overview of TAROT framework.} (top) Build a four-tier test suite (basic/intermediate/complex/edge) per problem using frontier LLMs and verify them against the reference solution.
(bottom) Reinforcement fine-tuning under a capability-conditioned, reward-decoupled curriculum. Less capable models perform best with basic $\rightarrow$ complex, whereas more capable models perform best with complex $\rightarrow$ basic.}
\label{fig:architecture}
\vspace{-6mm}
\end{figure*}

A coding problem $\mathcal{P}$ is formally defined as a tuple consisting of a problem statement $\mathcal{S}$, a reference solution $\mathcal{R}$, and a test suite $\mathcal{T}$:
\begin{align}
\mathcal{P} = (\mathcal{S}, \mathcal{R}, \mathcal{T}) \label{eq:problem_definition} 
\end{align}
In this structure, the problem statement $\mathcal{S}$ outlines the task, the reference solution $\mathcal{R}$ provides a correct implementation, and the test suite $\mathcal{T}$ serves for correctness verification without imposing a tiered difficulty structure.

From a software engineering perspective, development is commonly test-driven. It begins with simple tests and progressively incorporates more complex and edge cases, with implementations refactored along the way to improve both correctness and design. This staged expansion of tests naturally aligns with the intuition behind CL. However, typical coding problem test suites are not crafted with this incremental pedagogy in mind. They are primarily designed for summative verification, with highly variable size and difficulty, offering little support for CL.

To address this gap, we introduce the TAROT dataset \(\mathcal{D}_{\text{TAROT}}\), constructed according to the procedure depicted in Figure~\ref{fig:architecture} (a). Each problem is augmented with a tiered test suite spanning four predefined difficulty levels $L=\{\mathrm{B}, \mathrm{I}, \mathrm{C}, \mathrm{E}\}$ (basic, intermediate, complex, and edge), without modifying the original statement or the reference solution:
\begin{align}
\mathcal{D}_{\text{TAROT}}
&= \bigl\{\,\bigl(\mathcal{S}_i,\ \mathcal{R}_i,\ \{\,\mathcal{T}_{i,l}\,\}_{l\in L}\bigr)\ \bigr\}_{i=1}^{N},
\label{eq:tarot_dataset} \\
L &= \{\mathrm{B},\ \mathrm{I},\ \mathrm{C},\ \mathrm{E}\},
\label{eq:levels} \\
\mathcal{T}_i &= \bigcup_{l\in L}\ \mathcal{T}_{i,l},
\label{eq:tier_union} \\
\text{s.t.}\
\forall i \in[&N], \forall l\in L, \forall t\in \mathcal{T}_{i,l}: \Valid(\mathcal{R}_i, t).
\label{eq:verification}
\end{align}
The full suite for each problem is the union of its per-level subsets (Equation \eqref{eq:tier_union}) and every test case is validated against the reference solution to ensure data quality (Equation \eqref{eq:verification}). Any curriculum order (e.g., basic→complex) is imposed only during training and is not part of the dataset definition. 

\subsection{Training Mechanism} 
The training mechanism of TAROT decouples the curriculum from raw test scores. It achieves this by utilizing two pre-defined components: (1) a curriculum allocator that defines a fixed proportion of training focus for each difficulty tier $l \in L$, and (2) tailored reward weights that prioritize tiers by placing greater value where the learning signal is most beneficial. 

During the training loop, the model generates candidate solutions for a given problem, which are evaluated against the tiered test cases. The resulting pass/fail outcomes are used to calculate and accumulate a tier-weighted return. Both the curriculum allocation and reward weights are specified prior to training based on the model’s effective capability, a composite of instruction following fidelity and baseline coding proficiency. By conditioning these design choices on capability, TAROT concentrates the training signal on a model-specific zone of optimal difficulty, resulting in a fixed yet highly customized training schedule.
Specifically, models with lower baseline capability receive a larger share of basic and intermediate cases, whereas stronger models focus on complex and edge cases to extend their performance frontier. The reward weights follow the same principle, ensuring that successes on capability-appropriate tiers contribute more to the final objective.

Formally, in the RL setting, for each problem \(P_i\) and difficulty level \(l \in L\), we define the tier-level success of a policy \(\pi\) as the average pass rate over the corresponding test set:
\begin{equation}
r_{i,l}(\pi) \;=\; \frac{1}{|\mathcal{T}_{i,l}|}\sum_{t \in \mathcal{T}_{i,l}} \mathbf{1}\{\operatorname{Pass}(\pi, t)\},
\label{eq:per_level_success}
\end{equation}
where \(\operatorname{Pass}(\pi, t)\) indicates that the solution produced by \(\pi\) satisfies test case \(t\).

Given a curriculum allocation \(\boldsymbol{\alpha}=(\alpha_l)_{l\in L}\) and reward weights \(\boldsymbol{w}=(w_l)_{l\in L}\), the TAROT return for \(P_i\) is defined as:
\begin{equation}
\begin{aligned}
R_{\text{TAROT}}\!\big(P_i,\pi;\boldsymbol{\alpha},\boldsymbol{w}\big)
\;&=\; \sum_{l\in L} \alpha_l \, w_l \, r_{i,l}(\pi),\\
\sum_{l\in L}\alpha_l &= 1,\;\; w_l \ge 0.
\label{eq:tarot_return}
\end{aligned}
\end{equation}
Here, \(\alpha_l\) is the curriculum allocation that specifies the share of training updates to tier \(l\), and \(w_l\) is the reward weight that controls the contribution of tier \(l\)'s success to the overall return, reflecting its relevance given the model’s baseline capability.

Based on this tier-aggregated return, training is formulated as the maximization of the expected TAROT return over problems:
\begin{equation}
J_{\text{TAROT}}(\theta)\;=\;
\mathbb{E}_{P_i \sim \mathcal{D}_{\text{TAROT}}}
\Big[\, R_{\text{TAROT}}\!\big(P_i,\pi_\theta;\boldsymbol{\alpha},\boldsymbol{w}\big) \,\Big].
\label{eq:tarot_objective}
\end{equation}
By decoupling training effort allocation \(\boldsymbol{\alpha}\) from success valuation \(\boldsymbol{w}\), TAROT enables a capability-adaptive curriculum that concentrates optimization on the most informative difficulty tiers for a given model, rather than enforcing a fixed, model-agnostic learning path.

\section{Experiments}

\subsection{Experimental Settings}

\begin{figure*}[t!]
\centering
\includegraphics[width=\textwidth]{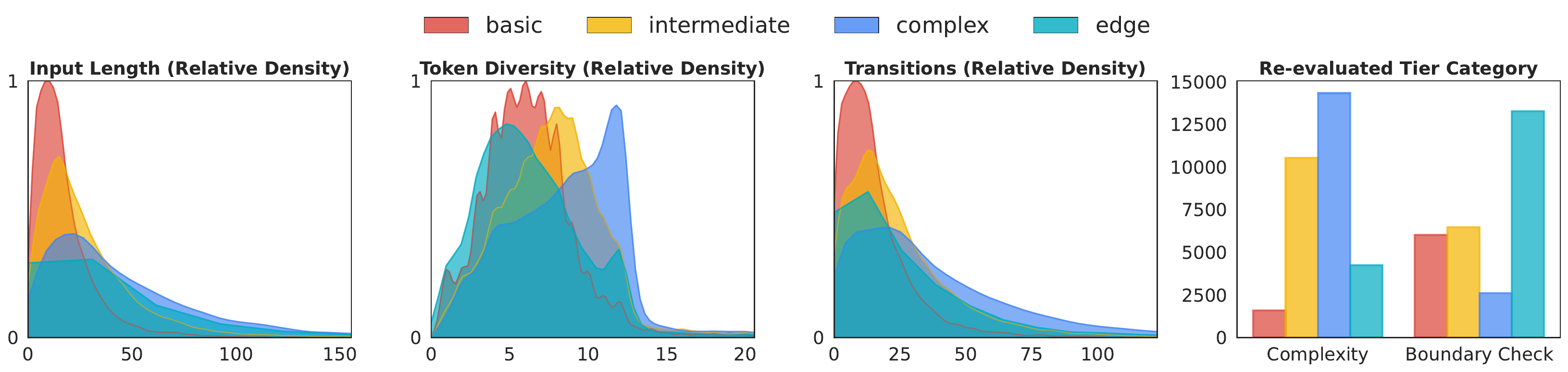}
\caption{Quantitative and qualitative validation of the TAROT dataset. The KDE plots show the distribution of structural complexity, where the x-axis represents the metric's magnitude. Token Diversity (unique/total tokens) and Transitions (character class changes) serve as proxies for lexical and syntactic complexity, respectively. The systematic rightward shift confirms increasing difficulty across tiers. GPT-4o validation on the right confirms that complex tiers target algorithmic complexity, while edge tiers focus on boundary conditions.}
\label{fig:tarot_dataset}
\vspace{-6mm}
\end{figure*}


\begin{table*}[t!]
\centering
\caption{Overview of the experimental schedules for curriculum learning. Each strategy varies in reward distribution and the sequence of difficulties presented to the model. The abbreviations B, I, C, and E correspond to basic, intermediate, complex, and edge difficulty tiers, respectively. For staged curricula, transitions occur at 0.2, 0.4, and 0.6 of the total epoch.}
\label{tab:scheduling_details}
\resizebox{0.85\textwidth}{!}{%
\begin{tabular}{@{}lcc@{}}
\toprule
\textbf{Strategy} & \textbf{Reward Weights (B, I, C, E)} & \textbf{Curriculum Schedule Progression} \\
\midrule

Forward (Uniform) & (0.25, 0.25, 0.25, 0.25) & B $\rightarrow$ (B,I) $\rightarrow$ (B,I,C) $\rightarrow$ All \\
\addlinespace

Forward (B \& I Weighted) & (0.35, 0.35, 0.15, 0.15) & B $\rightarrow$ (B,I) $\rightarrow$ (B,I,C) $\rightarrow$ All \\
\addlinespace

Forward (C \& E Weighted) & (0.15, 0.15, 0.35, 0.35) & B $\rightarrow$ (B,I) $\rightarrow$ (B,I,C) $\rightarrow$ All \\
\addlinespace

Reversed (C \& E Weighted) & (0.15, 0.15, 0.35, 0.35) & C $\rightarrow$ (C,E) $\rightarrow$ (C,E,I) $\rightarrow$ All \\
\addlinespace

Basic Only & (1.0, 0.0, 0.0, 0.0) & Static \\
\addlinespace

Complex Only & (0.0, 0.0, 1.0, 0.0) & Static \\
\addlinespace

Edge Only & (0.0, 0.0, 0.0, 1.0) & Static \\ 
\bottomrule
\end{tabular}%
}
\vspace{-2mm}
\end{table*}

We construct the TAROT dataset based on 15k Python coding interview problems\footnote{\url{https://huggingface.co/datasets/open-r1/verifiable-coding-problems-python}} with validated basic/intermediate/complex/edge test suites. As illustrated in Table ~\ref{tab:scheduling_details}, we design curriculum policies along two axes: allocation order and reward weighting. For allocation, we explore \textbf{Forward} (basic$\to$edge), \textbf{Reversed} (edge$\to$basic), and \textbf{Static} schedules. Transitions for staged curricula occur at 0.2, 0.4, and 0.6 of the total epoch. For weighting, we define three templates: \textbf{Uniform} ($0.25$ for all tiers), \textbf{B/I Weighted} (emphasizing the basic and intermediate tiers), and \textbf{C/Edge Weighted} (emphasizing the complex and edge tiers).

We first validate the empirical integrity of the TAROT dataset’s tiering by analyzing its structure using quantitative and qualitative metrics. Subsequently, we evaluate TAROT training mechanism on diverse LLMs, including Qwen2.5-Instruct, Qwen2.5-Coder-Instruct (1.5B, 3B, 7B) ~\citep{qwen2025qwen25technicalreport,hui2024qwen25codertechnicalreport}, Gemma-2-IT (2B, 9B)~\citep{gemmateam2024gemma2improvingopen}, and Qwen3-4B-Instruct-2507~\citep{yang2025qwen3technicalreport} on well-known benchmarks including HumanEval~\citep{chen2021codex}, MBPP~\citep{austin2021programsynthesislargelanguage}, HumanEval+, MBPP+~\citep{evalperf}, LiveCodeBench v5~\citep{jain2024livecodebenchholisticcontaminationfree}, CodeForces~\citep{penedo2025codeforces}, and CruxEval~\citep{gu2024cruxeval}. This selection allows us to assess the framework's effectiveness across a wide spectrum of model scales, architectures, coding specializations, and performance tiers. All models are fine-tuned using GRPO~\citep{shao2024deepseekmathpushinglimitsmathematical}. Full implementation details are presented in Appendix~\ref{app:training_details}.

\subsection{Main Results}
The structure analysis of the TAROT dataset is illustrated in Figure~\ref{fig:tarot_dataset}. The three KDE plots demonstrate a clear progression: as the tiers advance from basic to complex, the distributions for input length, token diversity, and character transitions all exhibit a consistent rightward shift, signifying a systematic increase in structural complexity. Furthermore, the qualitative bar chart reveals a functional separation between the two hardest tiers: test cases designed to probe complexity peak in the complex tier, while those targeting boundary checks are predominantly concentrated in the edge tier.
These results demonstrate that our four-level taxonomy not only stratifies overall difficulty but also effectively separates different types of challenge, establishing a robust foundation for subsequent experiments. 

\begin{figure*}[t]
\centering
\includegraphics[width=\textwidth]{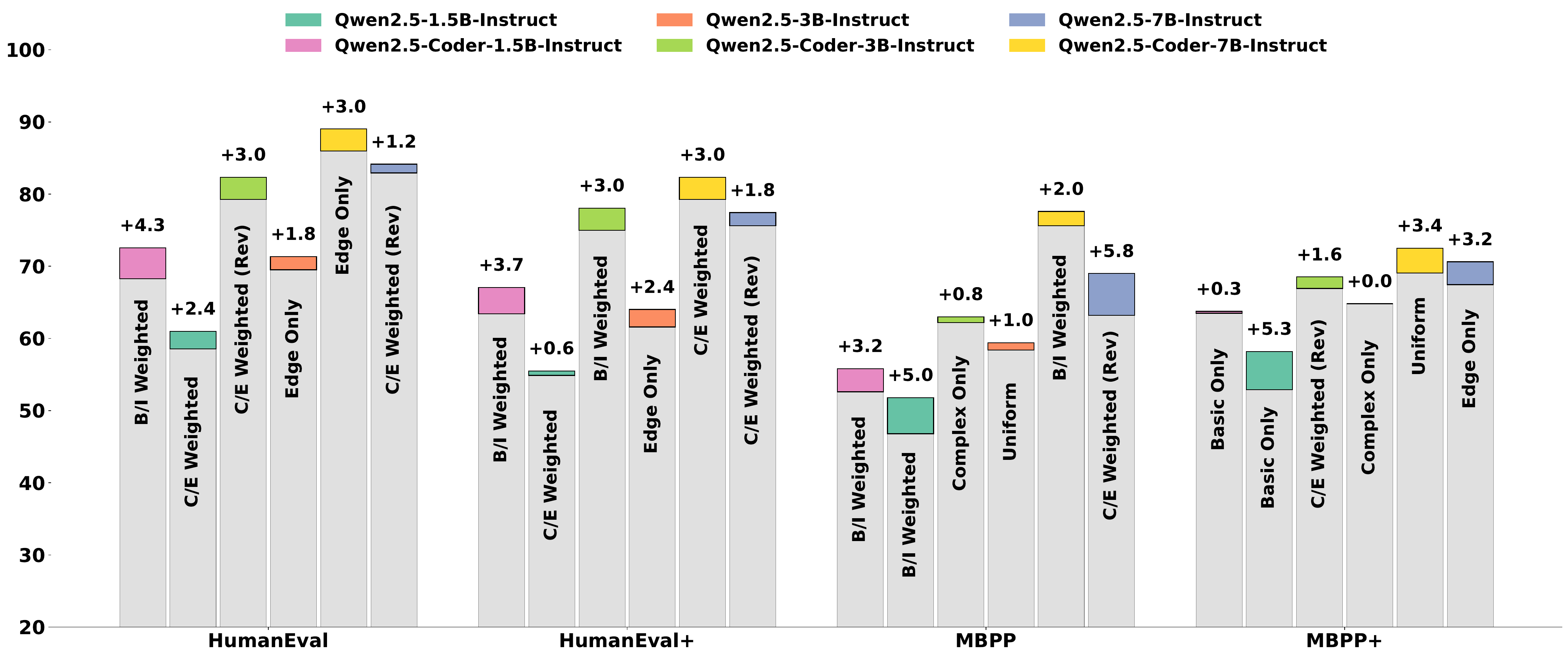}
\caption{Experimental results for Qwen2.5-Instruct and Qwen2.5-Coder-Instruct on HumanEval, HumanEval+, MBPP, and MBPP+. Scores are pass@1. Numbers above bars indicate gains in percentage points relative to each model’s base checkpoint. Labels inside bars indicate the best performing curriculum strategy.}
\label{fig:qwen25_instruct_winners}
\vspace{-6mm}
\end{figure*}

TAROT's training performances across various models on coding benchmarks are shown in Figure~\ref{fig:qwen25_instruct_winners}. Our approach consistently improves the pass@1 score over the base checkpoints across all benchmarks and model sizes, indicating its robustness and efficiency. 

These results reveal a nuanced relationship between model scale, specialization, and optimal curriculum design. 
The model scale influences the optimal curriculum design, which is exemplified by the trend shown in Qwen2.5-Instruct models. The largest model (7B) performs best with complex-focused strategies, while the smallest (1.5B) benefits from a conventional basic-focused approach. Model specialization further modulates the effective curriculum preference. As illustrated by the Qwen2.5-Coder models, the mid-scale Qwen2.5-Coder-3B model displays a learning preference akin to the much larger Instruct-7B model despite its smaller scale. It achieves its peak HumanEval score using the same complex-focused strategy and outperforms its general-purpose 3B counterpart. This suggests that a model's prior specialization enhances its effective capability, making it a more critical determinant of the ideal learning path than parameter count alone. 


We further evaluate TAROT on the more recent Qwen3-4B-Instruct-2507 to generalize these findings. As detailed in Table~\ref{tab:benchmarks-qwen3-comparison}, this model, fine-tuned with the optimal curriculum strategy \emph{C/E Weighted}, consistently outperforms the base model across all benchmarks, yielding gains ranging from +2.12 to +4.26 percentage points. Notably, these improvements are achieved on an already strong performance baseline, indicating that curriculum learning is effective even for eliciting further gains from highly capable models. 
In addition, these results align with our prior observations on the Qwen2.5 models.
They suggest that preference for curricula is primarily driven by a model’s effective capability rather than parameter count alone, and that stronger models benefit most from training regimes that prioritize challenging examples.

\begin{table*}[h!]
\centering
\caption{Performance comparison of curriculum strategies for Qwen3-4B-Instruct-2507. The best-performing \emph{C/E Weighted} strategy is compared against the base model, with improvements shown in percentage points.}
\label{tab:benchmarks-qwen3-comparison}
\resizebox{0.86\textwidth}{!}{%
\begin{tabular}{lcccc}
\toprule
\textbf{Strategy} & \textbf{HumanEval} & \textbf{HumanEval+} & \textbf{MBPP} & \textbf{MBPP+} \\
\midrule
Base & 89.02\% & 78.66\% & 52.60\% & 56.61\% \\
C/E Weighted & \textbf{91.46\%} (+2.44pp) & \textbf{82.92\%} (+4.26pp) & \textbf{55.20\%} (+2.60pp) & \textbf{58.73\%} (+2.12pp) \\
\bottomrule
\end{tabular}%
}
\vspace{-2mm}
\end{table*}

We attribute this divergence to the ``Zone of Optimal Difficulty''. For more-capable models, the basic tier is too trivial to provide informative learning signal, whereas the complex tier supplies the necessary high-entropy signal needed for improvement. Conversely, less-capable models facing complex tiers initially suffer from sparse rewards, leading to training collapse. Therefore, effective curriculum design must be calibrated to model capability: basic curricula for less-capable models to ensure training stability, while complex curricula for more-capable or code-specialized models to maximize gradient efficiency. 

\subsection{In-depth Analysis}

\begin{figure*}[t]
\centering
\includegraphics[width=\textwidth]{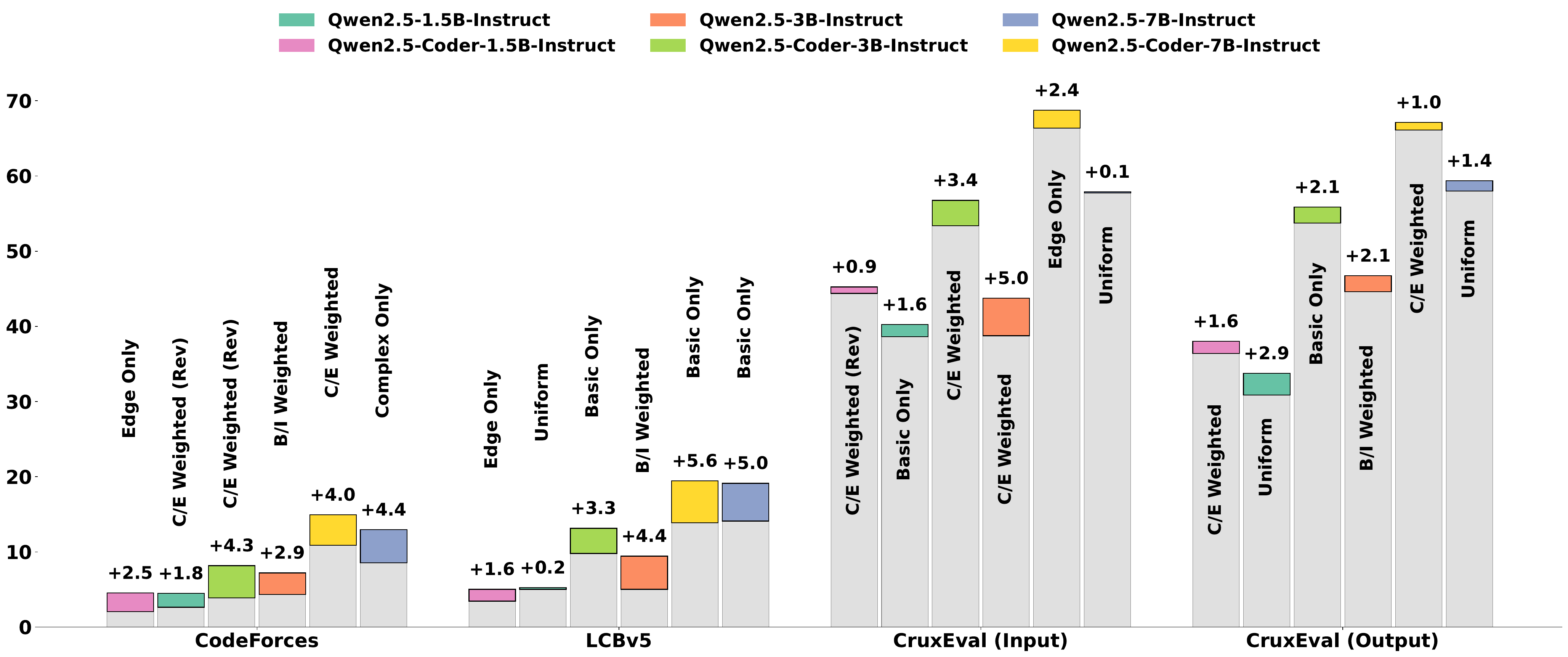}
\caption{Experimental results for Qwen2.5-Instruct and Qwen2.5-Coder-Instruct models on CodeForces, LiveCodeBench v5 (LCBv5), and CruxEval. Scores are the overall accuracy across easy, medium, and hard problems. Numbers above bars indicate gains in percentage points relative to each model’s base checkpoint. Labels inside bars indicate the best performing curriculum strategy.}
\label{fig:qwen25_instruct_winners_others}
\vspace{-3mm}
\end{figure*}

\paragraph{Out-Of-Distribution Benchmarks}
To assess the robustness of TAROT beyond the training distribution, we evaluate it on OOD benchmarks including CodeForces, LiveCodeBench v5, and CruxEval. Results are presented in Figure~\ref{fig:qwen25_instruct_winners_others}. We observe that TAROT consistently outperforms baselines; however, the optimal curriculum strategy is highly task-dependent rather than universal. 
For instance, Qwen2.5-7B achieved peak performance with the \emph{Basic Only} curriculum on LiveCodeBench v5, whereas the \emph{C/E Weighted} strategy proved most effective for CruxEval and CodeForces. This divergence stems from varying skill alignments between coding interview-style training data and downstream tasks. Consequently, effective curriculum selection must account for the target domain's computational structure, necessitating future research into task-specific intra-problem test design and adaptive policy selection.

\paragraph{Comparison with Standard Reward Schemes}
To verify the effectiveness of TAROT's capability-adaptive curriculum strategy, we compare our framework against standard reward strategies commonly used in RL for code generation. These baselines utilize the full four-tier test suite throughout training but do not incorporate curriculum scheduling. Specifically, we consider two standard RL baselines: 
\textbf{Avg-reward}, which computes rewards as the average pass rate across the four tiers ($R \in [0,1]$), and \textbf{Pass@All}, a strict functional correctness setting that assigns a reward of 1 only when all four tiers pass and 0 otherwise ($R \in \{0,1\}$). Results presented in Table ~\ref{tab:standard_baselines} show that TAROT consistently outperforms the standard reward schemes across all benchmarks, indicating that its performance gains arise from its capability-adaptive curriculum strategy.


\begin{table*}[t!]
\centering
\caption{Performance comparison between TAROT and standard RL Baselines. TAROT outperforms conventional reward schemes that use the same test cases but lack curriculum scheduling. Highest scores are highlighted in \textbf{bold}.}
\label{tab:standard_baselines}
\resizebox{\textwidth}{!}{%
\begin{tabular}{llccccccc}
\toprule
\textbf{Model} & \textbf{Strategy} & \textbf{HumanEval} & \textbf{HumanEval+} & \textbf{MBPP} & \textbf{MBPP+} & \textbf{CodeForces} & \textbf{LCBv5} & \textbf{CruxEval} \\
\midrule
\multirow{3}*{\textbf{Qwen2.5-1.5B-Instruct}}
 & Avg-reward & 59.15\% & 54.27\% & 49.20\% & 57.93\% & 2.72\% & 3.70\% & 38.75/32.87\% \\
 & Pass@All & \textbf{60.98\%} & \textbf{56.10\%} & 44.60\% & 53.43\% & 2.72\% & 3.94\% & 34.62/31.75\% \\
 & TAROT (Best) & \textbf{60.98\%} & 55.49\% & \textbf{51.80\%} & \textbf{58.20\%} & \textbf{4.49\%} & \textbf{5.26\%} & \textbf{40.20/33.75\%} \\ 
\midrule
\multirow{3}*{\textbf{Qwen2.5-7B-Instruct}}
 & Avg-reward & 83.75\% & 76.22\% & 66.00\% & 69.84\% & 11.41\% & 11.95\% & 56.37/58.50\% \\
 & Pass@All & 81.10\% & 73.78\% & 63.00\% & 68.52\% & 9.49\% & 14.81\% & 55.62/56.63\% \\
 & TAROT (Best) & \textbf{84.15\%} & \textbf{77.44\%} & \textbf{69.00\%} & \textbf{70.63\%} & \textbf{12.95\%} & \textbf{19.12\%} & \textbf{57.88/59.38\%} \\ 
\bottomrule
\end{tabular}
}
\end{table*}

\paragraph{Architectural Generalization}
We apply TAROT to the Gemma2 models, with results summarized in Table~\ref{tab:benchmarks-gemma-simplified}. Consistent with our findings on Qwen models, the optimal curriculum is governed by a model’s effective capability rather than its parameter count alone. For the larger Gemma2-9B-IT, \emph{Complex Only} curriculum offered no decisive advantage, while simpler strategies like \emph{Basic Only} often perform better on key benchmarks.
This principle is even more pronounced for the smaller Gemma2-2B-IT. As shown in Appendix~\ref{app:gemma2-2b-benchmarks}, most curricula are actively harmful, leading to a performance collapse from a sparse reward signal. In contrast, a \emph{Basic Only} strategy focuses on fundamentals yields substantial improvements. Together, these results indicate that for less-capable models, a fundamentals-first curriculum is essential for successful fine-tuning, whereas unstructured approaches can be severely detrimental.

\begin{table*}[h!]
\centering
\caption{Performance comparison for Gemma2-9B-IT across key curriculum strategies. Highest scores on each benchmark are highlighted in \textbf{bold}.}
\label{tab:benchmarks-gemma-simplified}
\resizebox{0.9\textwidth}{!}{%
\begin{tabular}{lccccccc}
\toprule
\textbf{Strategy} & \textbf{HumanEval} & \textbf{HumanEval+} & \textbf{MBPP} & \textbf{MBPP+} & \textbf{CodeForces} & \textbf{LCBv5} & \textbf{CruxEval} \\
\midrule
Base & 60.37\% & 54.88\% & 59.60\% & 65.08\% & 8.61\% & 11.83\% & 45.63/47.63\% \\
Uniform & \textbf{65.85}\% & 57.93\% & 59.20\% & 64.55\% & 10.82\% & 13.62\% & \textbf{51.63}/47.13\% \\
Basic Only & 63.41\% & 56.10\% & \textbf{60.40}\% & 65.08\% & 9.49\% & \textbf{14.70}\% & 51.00/48.00\% \\
Complex Only & \textbf{65.85}\% & \textbf{60.37}\% & 58.60\% & 64.55\% & 9.93\% & 12.54\% & 48.25/48.00\% \\
\bottomrule
\end{tabular}
}
\end{table*}

For completeness, we report the full per-strategy and per-curriculum results for
Qwen2.5-Instruct, Qwen2.5-Coder-Instruct, and Qwen3-4B-Instruct models across
HumanEval, MBPP, and the OOD benchmarks in
Appendix~\ref{app:qwen25_full_performance_table},
Table~\ref{tab:benchmarks-qwen3} and ~\ref{tab:benchmarks_qwen2.5}.
We further analyze the hyperparameter sensitivity (temperature, GRPO $\beta$)
and the inference-time maximum token limit in
Appendix~\ref{app:ablation_hyperparams} and ~\ref{app:ablation_max_tokens}, and investigate the training dynamics and reward correlations in Appendix~\ref{app:training_dynamics}.

\section{Conclusion}
We introduced TAROT, a test-driven and capability-adaptive framework for curriculum RFT in code generation. It moves beyond the conventional one-size-fits-all approach by constructing a four-tier, intra-problem test suite that allows curriculum design to be tailored to a model's effective capability. Extensive experiments validate TAROT's effectiveness in consistently improving model performance in code generation over strong baselines. In addition, we find that the optimal curriculum depends on a model’s effective capability rather than size alone. Specifically, we show that less-capable models benefit most from a basic-focused progression, while more-capable models excel with curricula that prioritize complex-focused challenges. These results and findings lay the groundwork for future research into automated and task-specific curriculum policies in code generation.

\section*{Limitations}
A primary limitation of our framework lies in its dependence on the TAROT dataset where the four-tier test suite is synthetically generated by frontier LLMs. Despite the rigorous verification process, potential biases or latent coverage gaps in the generator could propagate to the policy model and constrain the diversity of the learning signal. Additionally, the current study is restricted to Python coding tasks, and the generalization of this tiered reinforcement fine-tuning approach to multilingual or low-resource programming languages remains to be verified. Finally, while our capability-adaptive mechanism selects curriculum policies from a pre-defined portfolio based on static baseline assessments, we leave the exploration of continuous curriculum spaces for dynamic schedule optimization during training to future research.

\bibliography{ref}

\clearpage
\appendix
\section{Test Case Generation Prompts}
\label{app:generation_prompts}

To ensure the consistent generation of high-quality, four-tiered test cases, we designed a detailed prompt template. This template, listed in Table~\ref{tab:system_prompt}, guides the language model to act as an expert software engineer and produce test cases that adhere to our specific difficulty criteria.

\begin{table*}[h]
\caption{The prompt template used to generate a tiered test suite given a coding problem. The problem statement and the default test case from the original source are injected into \texttt{\{problem\_statement\}} and \texttt{\{baseline\_test\_case\}} placeholders, respectively.}
\label{tab:system_prompt}
\centering
\begin{tabular}{|p{\dimexpr\textwidth-2\tabcolsep-2\arrayrulewidth}|}
\hline\hline
\fontsize{7pt}{9.6pt}\selectfont 

You are an expert software engineer with extensive experience in designing comprehensive unit tests. Your task is to generate four distinct unit tests for a given code implementation based solely on the provided problem statement. Treat this as a black-box testing exercise---focus exclusively on the inputs and expected outputs without assuming any details about the internal implementation. 

\vspace{4pt} 

\textbf{Important:} A baseline test case will be provided separately. Each test case you generate must be more challenging than the baseline test case.

\vspace{4pt} 

Please generate four unit tests with the following guidelines:

\begin{enumerate}[leftmargin=*]
    \item \textbf{Basic Complexity Test (label as "basic"):}
    \begin{itemize}[leftmargin=*]
        \item Use simple, straightforward inputs.
        \item Validate the core behavior under normal conditions.
        \item Focus on the happy path scenario.
        \item This should be the least challenging test case relative to the others.
    \end{itemize}
    \item \textbf{Medium Complexity Test (label as "intermediate"):}
    \begin{itemize}[leftmargin=*]
        \item Include moderately complex inputs with some edge conditions.
        \item Test with mixed data types or larger inputs.
        \item Incorporate common edge cases and boundary values.
        \item Ensure this test is more challenging than the basic test.
    \end{itemize}
    \item \textbf{High Complexity Test (label as "complex"):}
    \begin{itemize}[leftmargin=*]
        \item Use complex, nested, or structured inputs.
        \item Validate advanced functionality and complex logic paths.
        \item Stress test the implementation with challenging scenarios.
        \item This test should be more intricate than both the basic and intermediate tests.
    \end{itemize}
    \item \textbf{Edge Case Test (label as "edge"):}
    \begin{itemize}[leftmargin=*]
        \item Use extreme boundary conditions and special cases.
        \item Validate behavior with empty, null, or invalid inputs.
        \item Focus on error handling and exception scenarios.
        \item This should be the most challenging test case among the four.
    \end{itemize}
\end{enumerate}

\vspace{4pt} 

For each test case, follow the JSON format provided in the example below (include only the input and expected output):

\begin{lstlisting}[breaklines=true, basicstyle=\ttfamily\tiny]
{
  "language": "python",
  "test_cases": [
    {
      "input": "4\n4\n0001\n1000\n0011\n0111\n3\n010\n101\n0\n2\n00000\n00001\n4\n01\n001\n0001\n00001\n",
      "output": "1\n3 \n-1\n0\n\n2\n1 2 \n",
      "type": "stdin_stdout",
      "label": "basic",
      "reason": "This test represents simple, straightforward input conditions."
    }
  ]
}
\end{lstlisting}

\vspace{4pt} 

Remember:
\begin{itemize}[leftmargin=*]
\item Do not assume any knowledge about the internal code; base your tests purely on the input-output behavior described in the problem statement. 
\item Ensure that each of your test cases is incrementally more challenging than the baseline test case provided.
\end{itemize} \\

\fontsize{7pt}{9.6pt}\selectfont \textbf{Problem Statement:} \texttt{\{problem\_statement\}} 

\vspace{4pt} 

\textbf{Baseline Test Case:} \texttt{\{baseline\_test\_case\}} \\
\hline\hline
\end{tabular}
\end{table*}

\section{Implementation Details}
\label{app:training_details}

\paragraph{TAROT Dataset}
Our experiments utilize the TAROT dataset, which is constructed by augmenting approximately 15k Python problems from the verifiable-coding-problems-python dataset\footnote{\url{https://huggingface.co/datasets/open-r1/verifiable-coding-problems-python}}. For each problem, we employ OpenAI's most powerful o3 and o4 models\footnote{\url{https://platform.openai.com/docs/models}} with the highest reasoning effort to generate a four-tiered test suite spanning basic, intermediate, complex, and edge cases. The prompts used for the generation are detailed in Table~\ref{tab:system_prompt}. To ensure high quality, every generated test case is validated using the reference solution, and any problem with even one failing tier is discarded. This rigorous curation process yields a final dataset consisting of 60k tiered test suites (15k problems × 4 tiers). Samples of the generated tiered test cases can be found in Appendix~\ref{app:sample_cases}.

\paragraph{Model Selection}
To validate our proposed framework, we select diverse models to investigate four key research questions: (1) the effect of model scale, to test whether the optimal curriculum is capability-dependent, using three Qwen2.5 models (1.5B, 3B, 7B)~\citep{qwen2025qwen25technicalreport}; (2) the impact of specialization, to determine if TAROT can further enhance models already proficient in coding, using their code-specialized counterparts~\citep{hui2024qwen25codertechnicalreport}; (3) architectural generalizability, to test if our findings apply beyond a single model family, by incorporating two instruction-tuned Gemma2 models (2B, 9B)~\citep{gemmateam2024gemma2improvingopen}; and (4) pushing performance frontiers, to assess if our framework can improve even state-of-the-art models with strong baselines, by fine-tuning the recent Qwen3-4B-Instruct-2507~\citep{yang2025qwen3technicalreport}. For all models, we use their instruction-tuned variants to ensure a foundational code-generation capability, a prerequisite for effective RL-based fine-tuning.


\paragraph{Training Details}
We fine-tune all selected models for a single epoch using the TAROT framework. For policy optimization, we employ GRPO~\citep{shao2024deepseekmathpushinglimitsmathematical}. All models are trained using the AdamW optimizer with a constant learning rate of $1 \times 10^{-6}$. We set the global batch size to 8, reducing it to 4 for larger models (Qwen2.5-7B-Instruct, Qwen2.5-Coder-7B-Instruct, and Gemma2-9B-IT) to accommodate memory constraints. The maximum input and completion token lengths are set to 1,024 and 4,096, respectively. For GRPO-specific settings, we generate 8 candidate completions per prompt to estimate the policy advantage, with the core hyperparameter $\beta$ set to 0.01 in our main experiments. The GRPO hyperparameter $\beta$ controls the strength of the Kullback-Leibler (KL) divergence regularization term, which penalizes the policy for deviating too far from the original base model's behavior. The training temperature, in turn, manages the exploration-exploitation trade-off; higher values encourage the model to sample a wider variety of solutions (exploration), while lower values cause it to refine high-probability ones (exploitation). To identify the optimal settings for the crucial hyperparameters, we provide an ablation study on the GRPO $\beta$ value (0.1, 0.05, 0.01) and the sampling temperature (1.0, 0.7, 0.5) in Appendix~\ref{app:ablation_hyperparams}. All fine-tuning experiments are conducted on a server with 8 x NVIDIA A100 (80 GB) GPUs, running CUDA 12.4 and PyTorch 2.6. Our implementation is based on open-source libraries including Transformers~\citep{wolf-etal-2020-transformers}, TRL~\citep{vonwerra2022trl}, vLLM~\citep{kwon2023efficient}, and Open-R1~\citep{openr1}.

\paragraph{Evaluation Metrics}
We evaluate the efficacy of TAROT on diverse code generation benchmarks. For functional correctness, we measure the pass@1 metric on HumanEval~\citep{chen2021codex}, MBPP~\citep{austin2021programsynthesislargelanguage}, HumanEval+, and MBPP+~\citep{evalperf}. To assess competitive problem-solving skills, we use the overall accuracy on LiveCodeBench v5~\citep{jain2024livecodebenchholisticcontaminationfree} and CodeForces~\citep{penedo2025codeforces}, averaged across their difficulty tiers. Finally, the model's code reasoning capability is evaluated using the input and output prediction accuracy on CruxEval~\citep{gu2024cruxeval}. The detailed generation parameters and execution environment are described in Appendix~\ref{app:exp_details}.

\section{Generation and Execution Environment}
\label{app:exp_details}

The entire evaluation pipeline is managed by the EvalChemy framework~\citep{evalchemy}. We follow the benchmark-specific generation configurations predefined within the framework, such as temperature, top-p, prompt formatting, and stopping criteria, to ensure consistency with established evaluation protocols. By default, the maximum completion tokens for each benchmark adhered to its standard setting; however, for an ablation study on generation length (Appendix~\ref{app:ablation_max_tokens}), we systematically increase this limit to 4{,}096, 8{,}192, and 16{,}384 tokens to observe performance trends.

All code generation for evaluation is conducted by serving the fine-tuned models via the vLLM framework~\citep{kwon2023efficient} on servers equipped with 4 x NVIDIA A100 (80 GB) GPUs, using a batch size of 64. The resulting code is executed in a secure, sandboxed Python 3.11 environment, where a strict 10-second timeout is enforced for each test case to prevent infinite loops and manage evaluation time.

\begin{figure*}[t]
\centering
\includegraphics[width=\textwidth]{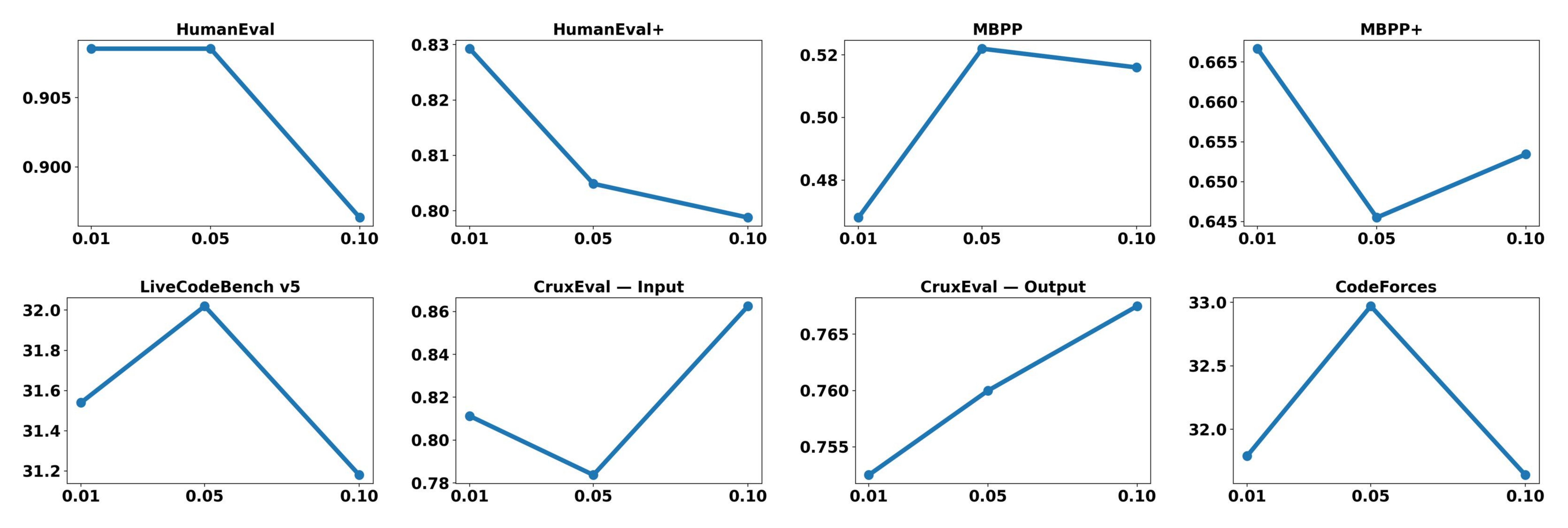}
\caption{Performance sensitivity to the GRPO hyperparameter $\beta$. The plots show the final pass@1 or accuracy scores on various benchmarks as $\beta$ is varied. The optimal value is task-dependent; for instance, HumanEval and HumanEval+ benefit from a smaller $\beta$ (0.01) that allows greater policy exploration, whereas MBPP and CodeForces achieve peak performance with a larger $\beta$ (0.05) that enforces stronger regularization.}
\label{fig:grpo_beta}
\end{figure*}

\begin{figure*}[t]
\centering
\includegraphics[width=\textwidth]{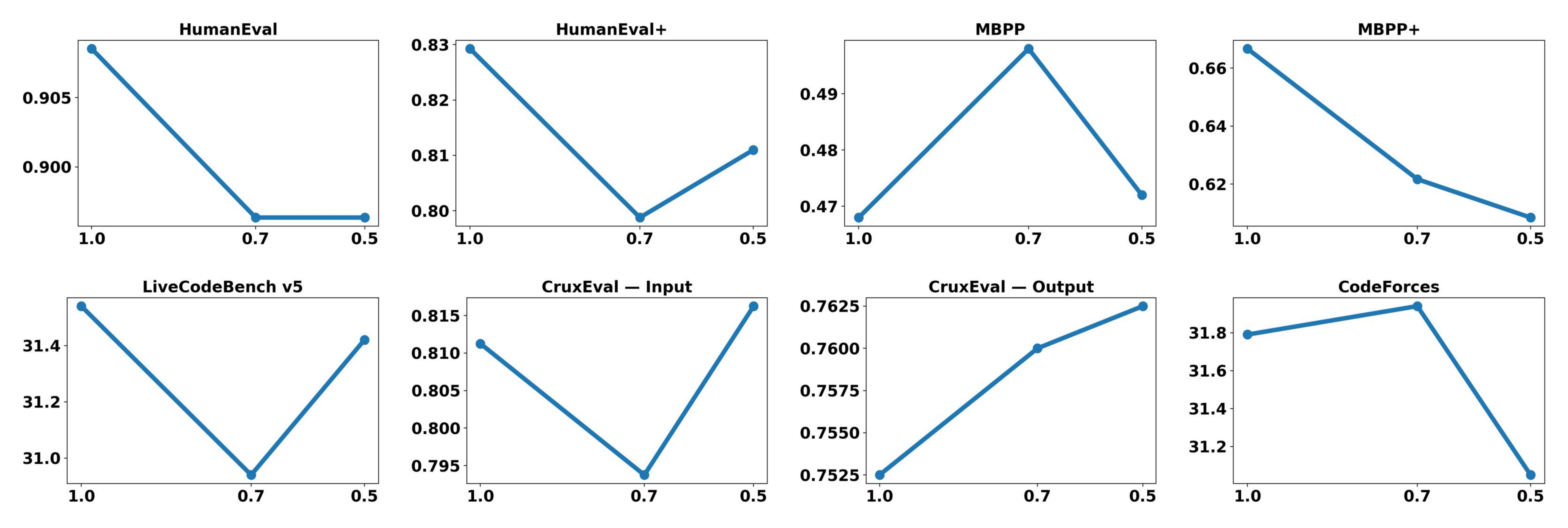}
\caption{Performance sensitivity to the sampling temperature during training. The plots illustrate the final benchmark scores for different training temperatures. A higher temperature of 1.0, which encourages greater exploration, is optimal for benchmarks like HumanEval and HumanEval+. In contrast, other benchmarks such as MBPP show a preference for a more moderate temperature of 0.7, highlighting that the ideal exploration-exploitation balance is task-specific.}
\label{fig:temperature}
\end{figure*}

\section{Hyperparameter Sensitivity Analysis}
\label{app:ablation_hyperparams}

This section provides ablation studies on two key training hyperparameters to analyze their impact on final benchmark performance: the GRPO regularization coefficient $\beta$ and the sampling temperature during training.

\paragraph{Impact of GRPO's $\beta$}
The hyperparameter $\beta$ in GRPO controls the strength of the Kullback-Leibler (KL) divergence regularization, which prevents the fine-tuned policy from deviating excessively from the original base model. The results of varying $\beta$ are shown in Figure~\ref{fig:grpo_beta}. Performance sensitivity to $\beta$ is not uniform across benchmarks. For function-synthesis tasks like HumanEval and HumanEval+, a small $\beta$ of 0.01, which allows for greater policy exploration, yields the best results. Conversely, benchmarks like MBPP and CodeForces appear to benefit from slightly stronger regularization ($\beta=0.05$). This variance suggests that the optimal regularization strength is task-dependent. We selected $\beta=0.01$ for our main experiments as it proved most effective on our primary evaluation benchmarks.

\paragraph{Impact of Training Temperature}
The sampling temperature manages the exploration-exploitation trade-off during training. The results, presented in Figure~\ref{fig:temperature}, indicate that a higher temperature of 1.0, which encourages greater exploration of diverse solutions, is optimal for HumanEval and HumanEval+. However, other benchmarks show different trends; MBPP, for example, peaks at a more conservative temperature of 0.7. This highlights that the optimal degree of exploration is also task-specific, and suggests that task-adaptive temperature scheduling could be a potential area for future work.

\section{Impact of Maximum Completion Tokens at Inference Time}
\label{app:ablation_max_tokens}

\begin{figure*}[t]
\centering
\includegraphics[width=\textwidth]{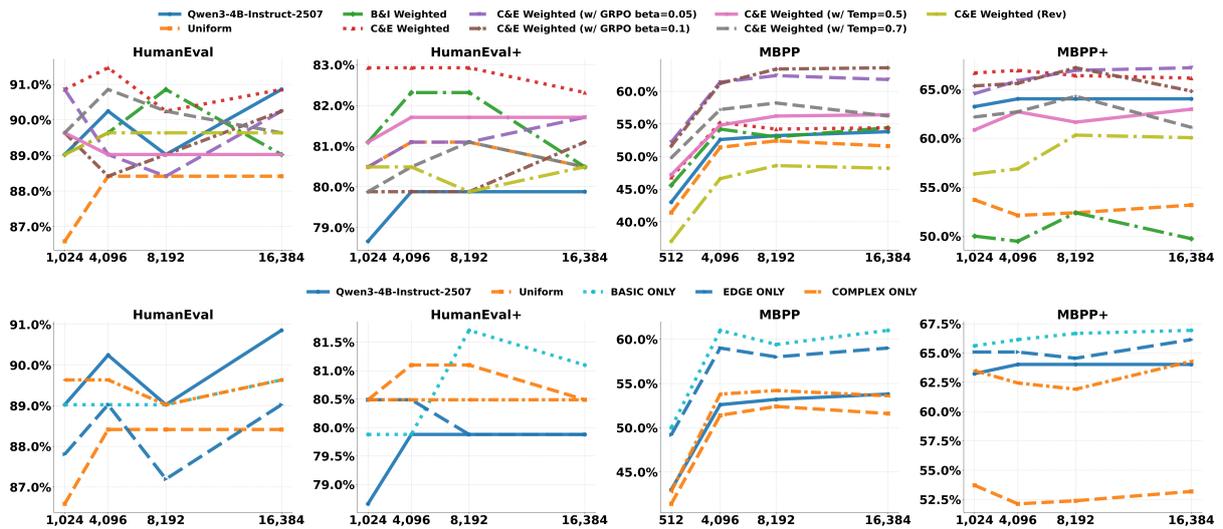}
\caption{Performance sensitivity to the maximum completion token limit at inference time for Qwen3-4B-Instruct-2507 fine-tuned on various curriculum strategies. The results reveal a clear, benchmark-dependent dichotomy. For function-completion tasks like HumanEval and HumanEval+, performance tends to degrade as the token limit increases beyond 4,096, suggesting that a larger generation space may encourage verbose, error-prone solutions. Conversely, for benchmarks like MBPP and MBPP+, a larger token limit is generally beneficial, indicating that their problem structures may require more extensive code to solve correctly.}
\label{fig:mt_trends}
\end{figure*}

We analyzed the impact of the maximum completion token limit during inference on the fine-tuned Qwen3-4B model, with results presented in Figure~\ref{fig:mt_trends}. The findings reveal a clear, benchmark-dependent dichotomy. On function-completion tasks like HumanEval and HumanEval+, performance generally degrades as the token limit increases beyond 4,096. In stark contrast, benchmarks like MBPP and MBPP+ benefit from a larger generation space, with optimal results often found at 8,192 or 16,384 tokens.

This divergence suggests that for tasks requiring concise solutions, such as those in HumanEval, a larger token limit may encourage verbose and error-prone code. Conversely, the nature of MBPP problems may necessitate a longer generation process to fully develop the correct logic. This analysis underscores a critical point for standardized evaluation: the ideal setting for maximum completion tokens is highly contingent on the characteristics of the target benchmark.

\section{Additional Results on Gemma2-2B-IT}
\label{app:gemma2-2b-benchmarks}

This appendix provides the full curriculum comparison for Gemma2-2B-IT as in Table~\ref{tab:benchmarks-gemma-2b-final}. Unlike larger or stronger models, Gemma2-2B-IT exhibits curriculum fragility: most curricula depress performance, consistent with the observation in the main text that sparse reward signals can cause collapse for less-capable models. In contrast, \emph{Basic Only}—a fundamentals-first schedule—yields the most reliable gains among the tested strategies.

These results reinforce our capability-dependent view of curriculum design: for weaker models, emphasizing simpler tiers is a prerequisite for successful fine-tuning, whereas complex-focused or mixed curricula can be harmful. 

\begin{table*}[h]
\centering
\caption{Performance comparison for Gemma2-2B-IT across all curriculum strategies. Scores are colored and bolded based on their deviation from the Base strategy (\textcolor{gaincolor}{\textbf{blue}} for higher, \textcolor{losscolor}{\textbf{red}} for lower).}
\label{tab:benchmarks-gemma-2b-final}
\resizebox{\textwidth}{!}{%
\begin{tabular}{lccccccc}
\toprule
\textbf{Strategy} & \textbf{HumanEval} & \textbf{HumanEval+} & \textbf{MBPP} & \textbf{MBPP+} & \textbf{CodeForces} & \textbf{LCBv5} & \textbf{CruxEval} \\
\midrule
Base & 42.07\% & 34.76\% & 41.20\% & 47.09\% & 2.21\% & 4.30\% & 37.50/26.88\% \\
Uniform & \textcolor{losscolor}{\textbf{39.02}}\% & \textcolor{losscolor}{\textbf{31.09}}\% & \textcolor{losscolor}{\textbf{33.80}}\% & \textcolor{losscolor}{\textbf{39.95}}\% & \textcolor{losscolor}{\textbf{0.22}}\% & \textcolor{losscolor}{\textbf{3.58}}\% & \textcolor{losscolor}{\textbf{33.00}}/\textcolor{losscolor}{\textbf{26.63}}\% \\
B/I Weighted & \textcolor{losscolor}{\textbf{35.98}}\% & \textcolor{losscolor}{\textbf{32.32}}\% & \textcolor{losscolor}{\textbf{35.60}}\% & \textcolor{losscolor}{\textbf{42.06}}\% & \textcolor{losscolor}{\textbf{0.22}}\% & 4.30\% & \textcolor{gaincolor}{\textbf{38.63}}/\textcolor{gaincolor}{\textbf{27.25}}\% \\
C/E Weighted & \textcolor{losscolor}{\textbf{41.46}}\% & \textcolor{losscolor}{\textbf{34.15}}\% & \textcolor{losscolor}{\textbf{39.20}}\% & \textcolor{gaincolor}{\textbf{48.41}}\% & \textcolor{losscolor}{\textbf{0.22}}\% & \textcolor{losscolor}{\textbf{3.94}}\% & \textcolor{losscolor}{\textbf{36.63}}/\textcolor{losscolor}{\textbf{26.75}}\% \\
C/E Weighted (Rev) & \textcolor{losscolor}{\textbf{40.86}}\% & \textcolor{gaincolor}{\textbf{35.37}}\% & \textcolor{losscolor}{\textbf{40.20}}\% & \textcolor{losscolor}{\textbf{44.44}}\% & \textcolor{losscolor}{\textbf{0.44}}\% & \textcolor{losscolor}{\textbf{3.94}}\% & \textcolor{losscolor}{\textbf{35.63}}/\textcolor{losscolor}{\textbf{26.75}}\% \\
Basic Only & \textcolor{gaincolor}{\textbf{44.51}}\% & \textcolor{gaincolor}{\textbf{37.20}}\% & \textcolor{losscolor}{\textbf{38.60}}\% & \textcolor{losscolor}{\textbf{46.83}}\% & \textcolor{losscolor}{\textbf{1.77}}\% & \textcolor{losscolor}{\textbf{3.94}}\% & \textcolor{gaincolor}{\textbf{39.88}}/\textcolor{gaincolor}{\textbf{27.88}}\% \\
Edge Only & \textcolor{losscolor}{\textbf{39.63}}\% & \textcolor{gaincolor}{\textbf{35.37}}\% & \textcolor{losscolor}{\textbf{38.00}}\% & \textcolor{losscolor}{\textbf{46.03}}\% & \textcolor{losscolor}{\textbf{0.22}}\% & 4.30\% & \textcolor{gaincolor}{\textbf{39.00}}/\textcolor{gaincolor}{\textbf{28.13}}\% \\
Complex Only & 42.07\% & \textcolor{gaincolor}{\textbf{36.59}}\% & \textcolor{losscolor}{\textbf{37.00}}\% & \textcolor{losscolor}{\textbf{45.77}}\% & 2.21\% & \textcolor{losscolor}{\textbf{2.87}}\% & \textcolor{losscolor}{\textbf{35.63}}/\textcolor{gaincolor}{\textbf{27.55}}\% \\
\bottomrule
\end{tabular}
}
\end{table*}

\section{Full Benchmark Tables (Qwen2.5 \& Qwen3-4B)}
\label{app:qwen25_full_performance_table}

\begin{table*}[b!]
\centering
\caption{Comprehensive performance evaluation of all curriculum strategies on Qwen3-4B-Instruct-2507. The highest score on each benchmark is highlighted in \textbf{bold}. The performance of Qwen3-Coder-30B-A3B-Instruct is included to enable comparison against a leading code-specialized model.}
\label{tab:benchmarks-qwen3}
\resizebox{\textwidth}{!}{%
\begin{tabular}{llccccccc}
\toprule
\textbf{Model} & \textbf{Strategy} & \textbf{HumanEval} & \textbf{HumanEval+} & \textbf{MBPP} & \textbf{MBPP+} & \textbf{CodeForces} & \textbf{LCBv5} & \textbf{CruxEval} \\
\midrule
\multicolumn{9}{l}{\textbf{Qwen3-Coder-30B-A3B-Instruct}} \\
 & Base & 94.51\% & 86.59\% & 73.80\% & 75.13\% & 29.65\% & 37.63\% & 81.75/79.25\% \\
\midrule
\multicolumn{9}{l}{\textbf{Qwen3-4B-Instruct-2507}} \\
 & Base & 89.02\% & 78.66\% & 52.60\% & 56.61\% & 33.63\% & 32.02\% & 78.25/\textbf{77.75\%} \\
 & Uniform & 88.41\% & 80.09\% & 35.30\% & 53.70\% & 31.86\% & 31.96\% & 79.37/75.75\% \\ 
 & B/I Weighted & 89.63\% & 81.09\% & 28.00\% & 52.38\% & 33.04\% & \textbf{33.81\%} & 79.50/75.38\% \\ 
 & C/E Weighted & \textbf{91.46\%} & \textbf{82.92\%} & \textbf{55.20\%} & \textbf{58.73\%} & 31.79\% & 31.54\% & \textbf{81.12}/75.25\% \\ 
 & C/E Weighted (Rev) & 89.63\% & 80.48\% & 36.20\% & 35.98\% & \textbf{34.66\%} & 31.66\% & 79.50/76.00\% \\ 
 & Basic Only & 89.63\% & 79.87\% & 39.80\% & 56.34\% & 33.11\% & 31.90\% & 78.50/75.00\% \\ 
 & Edge Only & 89.63\% & 79.88\% & 47.20\% & 56.61\% & 31.86\% & 30.59\% & 80.25/74.00\% \\ 
 & Complex Only & 90.85\% & 80.48\% & 28.60\% & 51.85\% & 30.61\% & 31.30\% & 80.37/76.37\% \\ 
\bottomrule
\end{tabular}
}
\end{table*}

We report the complete benchmark results for all curriculum strategies on Qwen2.5 family (1.5B/3B/7B, including Coder variants) and Qwen3-4B-Instruct-2507. These tables expand the main figures by listing pass@1 on HumanEval, HumanEval+, MBPP, MBPP+, and average accuracy of CodeForces, LiveCodeBench v5, and CruxEval for every strategy in Table~\ref{tab:benchmarks_qwen2.5} and~\ref{tab:benchmarks-qwen3}.

Consistent with the main text, the \emph{C/E Weighted} strategy tends to be the top performer for the more-capable Qwen3-4B model, improving over the base across all four code-function benchmarks. The full per-strategy breakdowns here allow exact comparison across OOD benchmarks as well.

\begin{table*}[]
\centering
\caption{Comprehensive performance evaluation of all curriculum strategies across Qwen2.5-Instruct and Qwen2.5-Coder-Instruct models (1.5B, 3B, 7B). For each model size, the highest score on each benchmark is highlighted in \textbf{bold}.}
\label{tab:benchmarks_qwen2.5}
\resizebox{\textwidth}{!}{%
\begin{tabular}{llccccccc}
\toprule
\textbf{Model} & \textbf{Strategy} & \textbf{HumanEval} & \textbf{HumanEval+} & \textbf{MBPP} & \textbf{MBPP+} & \textbf{CodeForces} & \textbf{LCBv5} & \textbf{CruxEval} \\
\midrule
\multicolumn{9}{l}{\textbf{Qwen2.5-7B-Instruct}} \\
 & Base & 82.93\% & 75.61\% & 63.20\% & 67.46\% & 8.54\% & 14.10\% & 57.75/58.00\% \\
 & Uniform & 82.93\% & 73.78\% & 67.40\% & 67.46\% & 12.36\% & 14.93\% & \textbf{57.88}/\textbf{59.38\%} \\
 & B/I Weighted & 78.05\% & 76.83\% & 67.60\% & 69.58\% & 11.56\% & 15.89\% & 57.25/59.25\% \\
 & C/E Weighted & 79.27\% & 73.78\% & 66.20\% & 70.37\% & 10.89\% & 15.77\% & 57.13/55.50\% \\
 & C/E Weighted (Rev) & \textbf{84.15\%} & \textbf{77.44\%} & \textbf{69.00\%} & 70.11\% & 8.24\% & 15.41\% & \textbf{57.88}/56.38\% \\
 & Basic Only & 82.32\% & 75.61\% & 66.20\% & 68.52\% & 12.29\% & \textbf{19.12\%} & 55.63/57.50\% \\
 & Edge Only & 83.54\% & 76.22\% & 67.60\% & \textbf{70.63\%} & 11.11\% & 17.08\% & 56.13/57.75\% \\
 & Complex Only & \textbf{84.15\%} & 75.61\% & \textbf{69.00\%} & 69.05\% & \textbf{12.95\%} & 17.80\% & 57.25/56.38\% \\
\midrule
\multicolumn{9}{l}{\textbf{Qwen2.5-3B-Instruct}} \\
 & Base & 69.51\% & 61.59\% & 58.40\% & \textbf{64.81\%} & 4.34\% & 5.02\% & 38.75/44.63\% \\
 & Uniform & \textbf{71.34\%} & 63.41\% & \textbf{59.40\%} & 63.49\% & 6.92\% & 8.72\% & 42.00/42.50\% \\
 & B/I Weighted & 69.51\% & 62.20\% & 59.00\% & 63.49\% & \textbf{7.21\%} & \textbf{9.44\%} & 42.38/\textbf{46.75\%} \\
 & C/E Weighted & 69.51\% & 62.80\% & 56.60\% & 63.76\% & \textbf{7.21\%} & 7.17\% & \textbf{43.75}/44.50\% \\
 & C/E Weighted (Rev) & 70.12\% & 62.80\% & 57.00\% & 63.49\% & 6.92\% & 8.00\% & 43.63/42.50\% \\
 & Basic Only & 66.46\% & 59.15\% & \textbf{59.40\%} & 64.02\% & 6.33\% & 6.09\% & 40.50/44.13\% \\
 & Edge Only & \textbf{71.34\%} & \textbf{64.02\%} & 58.20\% & 62.70\% & 6.11\% & 7.05\% & 43.13/42.63\% \\
 & Complex Only & 67.68\% & 60.37\% & 59.00\% & \textbf{64.81\%} & 6.84\% & 6.33\% & 41.25/42.88\% \\
\midrule
\multicolumn{9}{l}{\textbf{Qwen2.5-1.5B-Instruct}} \\
 & Base & 58.54\% & 54.88\% & 46.80\% & 52.91\% & 2.65\% & 5.02\% & 38.63/30.88\% \\
 & Uniform & \textbf{60.98\%} & 54.88\% & 50.00\% & 57.14\% & 3.68\% & \textbf{5.26\%} & 37.13/\textbf{33.75\%} \\
 & B/I Weighted & 59.15\% & 54.27\% & \textbf{51.80\%} & 57.94\% & 3.83\% & 4.54\% & 36.00/29.75\% \\
 & C/E Weighted & \textbf{60.98\%} & \textbf{55.49\%} & 49.40\% & 56.08\% & 3.61\% & 5.02\% & 34.75/32.38\% \\
 & C/E Weighted (Rev) & 56.71\% & 52.44\% & 50.40\% & \textbf{58.20\%} & \textbf{4.49\%} & 4.90\% & 34.00/31.75\% \\
 & Basic Only & 57.32\% & 53.05\% & 50.60\% & \textbf{58.20\%} & 4.05\% & 4.66\% & \textbf{40.25}/33.00\% \\
 & Edge Only & 55.49\% & 50.61\% & 50.20\% & 56.08\% & 3.75\% & 4.42\% & 35.50/31.50\% \\
 & Complex Only & 59.76\% & 54.88\% & \textbf{51.80\%} & 55.29\% & 3.46\% & 4.54\% & 36.13/33.38\% \\
\midrule
\midrule
\multicolumn{9}{l}{\textbf{Qwen2.5-Coder-7B-Instruct}} \\
 & Base & 85.98\% & 79.27\% & 75.60\% & 69.05\% & 10.89\% & 13.86\% & 66.38/66.13\% \\
 & Uniform & 85.76\% & 79.27\% & 77.20\% & \textbf{72.49\%} & 13.98\% & 17.68\% & 66.50/66.38\% \\
 & B/I Weighted & 84.76\% & 78.66\% & \textbf{77.60\%} & 71.96\% & 13.32\% & 17.44\% & 68.38/65.88\% \\
 & C/E Weighted & 87.80\% & \textbf{82.32\%} & 76.20\% & 70.90\% & \textbf{14.94\%} & 19.24\% & 66.25/\textbf{67.13\%} \\
 & C/E Weighted (Rev) & \textbf{88.41\%} & 81.10\% & 75.00\% & 71.42\% & 13.98\% & 19.12\% & 68.63/65.00\% \\
 & Basic Only & 85.98\% & 79.88\% & 76.20\% & 71.96\% & 14.86\% & \textbf{19.47\%} & 67.50/66.38\% \\
 & Edge Only & 79.02\% & 81.07\% & 77.20\% & 71.96\% & 12.14\% & 19.12\% & \textbf{68.75}/66.00\% \\ 
 & Complex Only & 87.80\% & 80.49\% & 76.60\% & 70.90\% & 14.35\% & 18.16\% & 67.75/66.50\% \\
\midrule
\multicolumn{9}{l}{\textbf{Qwen2.5-Coder-3B-Instruct}} \\
 & Base & 79.27\% & 75.00\% & 62.20\% & 66.93\% & 3.90\% & 9.80\% & 53.38/53.75\% \\
 & Uniform & 81.10\% & 76.83\% & 62.00\% & 67.20\% & 7.21\% & 10.75\% & 54.00/54.75\% \\
 & B/I Weighted & 81.71\% & \textbf{78.05\%} & 61.40\% & 66.93\% & 6.70\% & 9.80\% & 54.25/53.38\% \\
 & C/E Weighted & 79.88\% & 76.83\% & 61.00\% & 67.46\% & 8.02\% & 10.51\% & \textbf{56.75}/53.50\% \\
 & C/E Weighted (Rev) & \textbf{82.32\%} & 77.44\% & 62.00\% & \textbf{68.52\%} & \textbf{8.17\%} & 10.75\% & 52.63/55.13\% \\
 & Basic Only & 80.49\% & 76.22\% & 62.80\% & 66.67\% & 7.95\% & \textbf{13.14\%} & 55.88/\textbf{55.88\%} \\
 & Edge Only & 79.27\% & 75.00\% & 62.60\% & 66.14\% & 7.21\% & 10.63\% & 53.75/53.25\% \\
 & Complex Only & 78.05\% & 73.78\% & \textbf{63.00\%} & 67.72\% & 7.65\% & 10.63\% & 53.13/55.25\% \\
\midrule
\multicolumn{9}{l}{\textbf{Qwen2.5-Coder-1.5B-Instruct}} \\
 & Base & 68.29\% & 63.41\% & 52.60\% & 63.49\% & 2.06\% & 3.46\% & 44.38/36.38\% \\
 & Uniform & 71.34\% & 65.24\% & 52.80\% & 62.96\% & 4.56\% & 4.42\% & 44.75/36.38\% \\
 & B/I Weighted & \textbf{72.56\%} & \textbf{64.02\%} & \textbf{55.80\%} & 62.70\% & 4.19\% & 4.66\% & 45.13/35.75\% \\
 & C/E Weighted & 71.34\% & 66.65\% & 54.60\% & 62.96\% & 3.46\% & 4.42\% & 45.13/\textbf{38.00\%} \\
 & C/E Weighted (Rev) & \textbf{72.56\%} & 64.20\% & 54.20\% & 62.96\% & 3.38\% & 4.18\% & \textbf{45.25}/37.00\% \\
 & Basic Only & 70.12\% & 64.02\% & 54.00\% & \textbf{64.76\%} & 4.49\% & 4.66\% & 43.25/36.00\% \\
 & Edge Only & \textbf{72.56\%} & \textbf{67.10\%} & 53.60\% & 62.17\% & \textbf{4.56\%} & \textbf{5.02\%} & 44.86/35.63\% \\
 & Complex Only & 71.34\% & 66.46\% & 53.20\% & 63.49\% & 3.31\% & 4.54\% & 43.75/37.36\% \\
\bottomrule
\end{tabular}
}
\end{table*}

\section{Training Dynamics Analysis}
\label{app:training_dynamics}

\begin{figure*}[t]
\centering
\includegraphics[width=\textwidth]{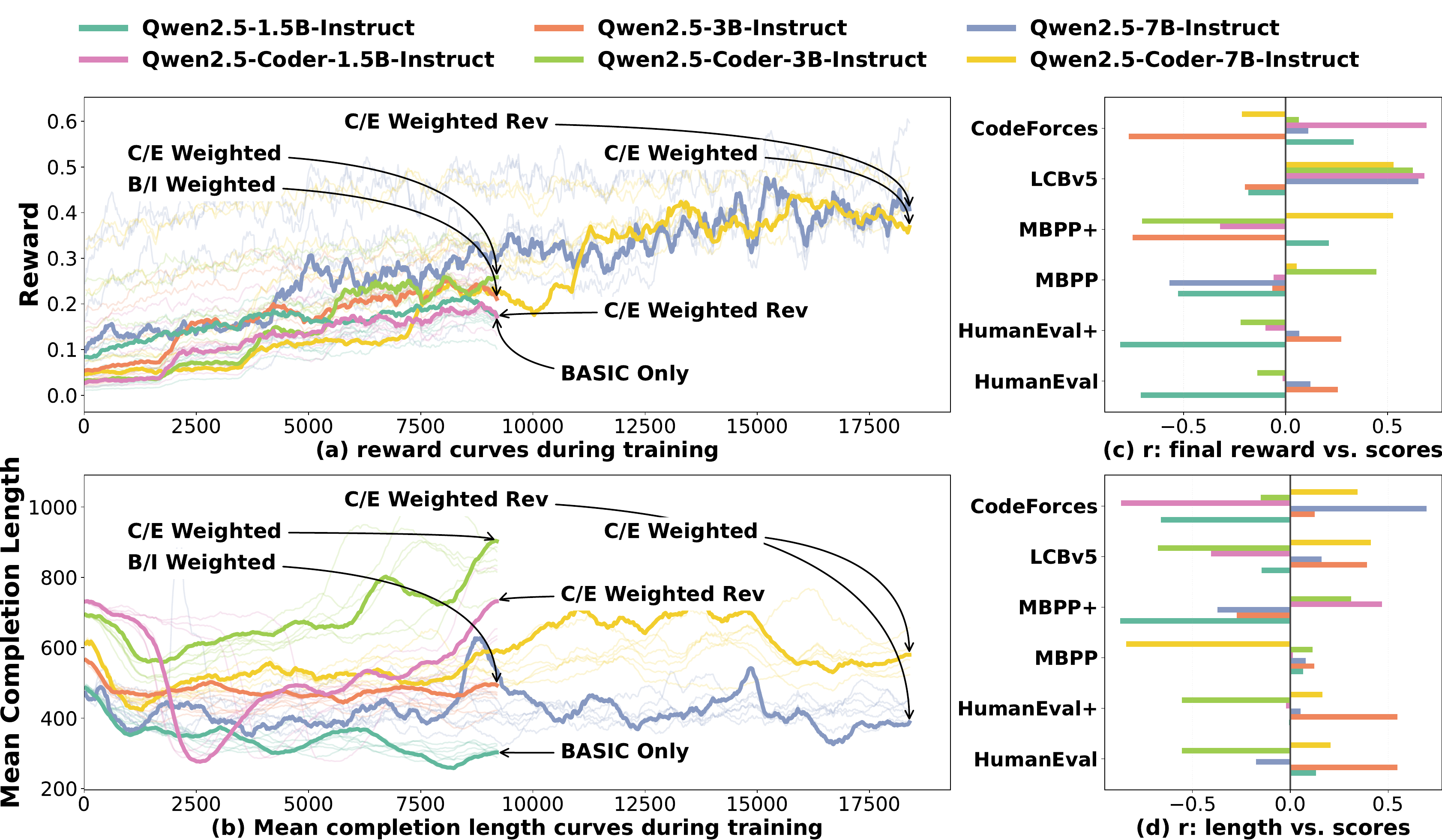}
\caption{Training dynamics vs. downstream performance. (a) and (b) show the reward and the mean completion length curves during reinforcement fine-tuning, and the annotations mark the curriculum strategy with the best average downstream performance. (c) and (d) show the Pearson correlation coefficient $r$ of the final rewards vs. benchmark scores and the mean completion length vs. benchmark scores, respectively. Light, semi-transparent lines represent alternative curriculum strategies, while the solid, annotated lines correspond to the best-performing strategy for each model. Some trajectories terminate earlier than others because different model sizes utilize varying batch sizes and gradient accumulation steps under a fixed total compute budget.}
\label{fig:reward-trends}
\end{figure*}

\paragraph{The Limits of the Reward Signal.}
Figure~\ref{fig:reward-trends} (a) shows that the training reward increases stably and is clearly separated by model capacity, indicating a stable optimization process. Note that initial rewards are relatively low even for capable models; this is due to strict output formatting requirements and execution timeouts enforced by the sandbox, which the models quickly adapt to during the early stages of fine-tuning.
This pattern suggests that the policy learns the training distribution well and that stronger models achieve higher reward levels under the same curriculum. 
However, the reward observed during training does not reliably anticipate downstream benchmark outcomes. 
As shown in Figure~\ref{fig:reward-trends} (c), the final reward has only a weak Pearson correlation coefficient with benchmark scores, which means that runs with similar rewards can still deliver very different levels of task performance. 

\paragraph{Conciseness as a Proxy for Advanced Reasoning.}
A different perspective comes from analyzing completion length.
Figure~\ref{fig:reward-trends} (b) shows that models with greater capability tend to produce shorter solutions as training progresses, and this tendency becomes more pronounced for stronger configurations.
Importantly, Figure~\ref{fig:reward-trends} (d) indicates that mean completion length exhibits a stronger negative correlation with benchmark scores than the reward does, implying that conciseness aligns better with final solution quality.
Shorter programs are more likely to capture the essential reasoning steps without unnecessary detours, whereas longer outputs often reflect uncertainty or inefficient search.
These observations support using solution conciseness as a practical secondary proxy for advanced reasoning quality, complementing the reward based perspective and providing a more informative early indicator of downstream performance.

\section{The Use of AI Assistants}
AI Assistants (\textit{e.g.,} LLMs) are employed solely for polishing writing.

\section{Sample Tiered Test Cases}
\label{app:sample_cases}

Table~\ref{tab:sample1}-~\ref{tab:sample7} present concrete examples of the four-tiered test cases generated for several problems in the TAROT dataset. These samples illustrate a clear and intentional progression in difficulty and scope, which is a cornerstone of our framework.

The tiers are generally designed to validate different aspects of a solution. Basic tiers focus on the core logic of a problem with simple, straightforward inputs. Following this, intermediate and complex tiers introduce greater difficulty through larger inputs, more intricate scenarios, or patterns requiring more sophisticated algorithmic reasoning. Finally, edge tiers are designed to test for robustness by probing boundary conditions, constraints, and performance-intensive cases such as large numbers or long strings. This tiered structure exemplifies the intra-problem difficulty gradient that forms the basis of our capability-adaptive curriculum.

\clearpage
\begin{table*}[t!]
\vspace*{-0.4cm}
\fontsize{8}{10}\selectfont
\centering
\caption{A sample from TAROT dataset comprising 4-tiered test cases: basic, intermediate, complex, and edge. The Reason column details the rationale for each tier assignment.}
\label{tab:sample1}
\begin{tabularx}{\textwidth}{@{} m{1.5cm} Y Y Y Y @{}}
\toprule
\multicolumn{5}{>{\hsize=\dimexpr5\hsize+8\tabcolsep\relax}Y}{
\begin{minipage}[c]{\hsize} 
\begin{alltt}
Solve the following coding problem using the programming language python:


You are given two positive integer numbers a and b. Permute (change order) of the digits of a to construct maximal number not exceeding b. No number in input and/or output can start with the digit 0. 


It is allowed to leave a as it is.


Input

The first line contains integer a ($1 \le a \le 10^{18}$). The second line contains integer b ($1 \le b \le 10^{18}$). Numbers don't have leading zeroes. It is guaranteed that answer exists.


Output

Print the maximum possible number that is a permutation of digits of a and is not greater than b. The answer can't have any leading zeroes. It is guaranteed that the answer exists. The number in the output should have exactly the same length as number a. It should be a permutation of digits of a.


Examples

Input

123
222


Output

213



Input

3921
10000


Output

9321



Input

4940
5000


Output

4940


The input will be given via stdin and the output should be printed to stdout by your code.


Now solve the problem by providing the code.
\end{alltt}

\end{minipage} %
} \\
\midrule
\textbf{\small Test cases} & \textbf{\small Basic} & \textbf{\small Intermediate} & \textbf{\small Complex} & \textbf{\small Edge} \\
\midrule
\textbf{\small Input} & 
\texttt{21} \par \texttt{12} & 
\texttt{3051} \par \texttt{5310} & 
\texttt{98761230} \par \texttt{98765000} & 
\texttt{111222333444555666} \par \texttt{1000000000000000000} \\
\addlinespace 
\textbf{\small Output} & 
\texttt{12} & 
\texttt{5310} & 
\texttt{98763210} & 
\texttt{666555444333222111} \\
\addlinespace 
\textbf{\small Reason} & 
A simple 2-digit case where swapping the digits yields the only valid permutation $\le$ bound, illustrating the happy path. & 
A 4-digit case including zero, requiring the algorithm to match the upper bound exactly with a permutation of the digits. & 
An 8-digit case where matching the bound fails at a later position, forcing backtracking and a maximal tail fill. & 
An extreme boundary case with an 18‑digit input and a longer 19‑digit bound, where any valid permutation fits, so the result is the digits sorted in descending order. \\
\bottomrule
\end{tabularx}
\end{table*}

\clearpage
\begin{table*}[t!]
\vspace*{-0.4cm}
\fontsize{8}{10}\selectfont
\centering
\caption{A sample from TAROT dataset comprising 4-tiered test cases: basic, intermediate, complex, and edge. The Reason column details the rationale for each tier assignment.}
\label{tab:sample2}
\begin{tabularx}{\textwidth}{@{} m{1.5cm} Y Y Y Y @{}}
\toprule
\multicolumn{5}{>{\hsize=\dimexpr5\hsize+8\tabcolsep\relax}Y}{
\begin{minipage}[c]{\hsize} 
\begin{alltt}
Solve the following coding problem using the programming language python:


Winter is here at the North and the White Walkers are close. John Snow has an army consisting of n soldiers. While the rest of the world is fighting for the Iron Throne, he is going to get ready for the attack of the White Walkers.


He has created a method to know how strong his army is. Let the $i$-th soldier's strength be $a_{i}$. For some $k$, we call the indices $i_1, i_2, \dots, i_k$ a clan if $i_1 < i_2 < \dots < i_k$ and $\gcd(a_{i_1}, a_{i_2}, \dots, a_{i_k}) > 1$. The strength of that clan is defined as $k\cdot\gcd(a_{i_1}, a_{i_2}, \dots, a_{i_k})$. The strength of the army is defined by the sum of the strengths of all possible clans.


Your task is to find the strength of his army. As the number may be very large, you have to print it modulo 1000000007 ($10^{9} + 7$).


Greatest common divisor ($\gcd$) of a sequence of integers is the maximum possible integer so that each element of the sequence is divisible by it.


-----Input-----

The first line contains integer n $(1 \leq n \leq 200000)$ $ - $ the size of the army. The second line contains n integers $a_1, a_2, ..., a_n$ $(1 \leq a_{i} \leq 1000000)$ $ - $ denoting the strengths of his soldiers.


-----Output-----

Print one integer $ - $ the strength of John Snow's army modulo 1000000007 ($10^{9} + 7$).


-----Examples-----

Input

3

3 3 1


Output

12



Input

4

2 3 4 6


Output

39


-----Note-----

In the first sample the clans are $\{1\}, \{2\}, \{1, 2\}$ so the answer will be $1\cdot3 + 1\cdot3 + 2\cdot3 = 12$


The input will be stdin and you should print your solution to stdout


Now solve the problem and return the code.


\end{alltt}

\end{minipage} %
} \\
\midrule
\textbf{\small Test cases} & \textbf{\small Basic} & \textbf{\small Intermediate} & \textbf{\small Complex} & \textbf{\small Edge} \\
\midrule
\textbf{\small Input} & 
\texttt{4} \par \texttt{2 3 5 7} & 
\texttt{6} \par \texttt{2 4 8 3 9 6} & 
\texttt{7} \par \texttt{2 2 2 2 2 2 2} & 
\texttt{5} \par \texttt{1 1 1 1 1} \\
\addlinespace 
\textbf{\small Output} & 
\texttt{17} & 
\texttt{119} & 
\texttt{896} & 
\texttt{0} \\
\addlinespace 
\textbf{\small Reason} & 
All strengths are prime, so only single‐soldier clans contribute. & 
Mix of primes and composites yields clans of various sizes and gcds. & 
Uniform strengths where every nonempty subset is a valid clan (gcd=2). & 
All strengths are 1, so no clan has gcd>1; result is zero. \\
\bottomrule
\end{tabularx}
\end{table*}

\clearpage
\begin{table*}[t!]
\vspace*{-0.4cm}
\fontsize{7}{7.5}\selectfont
\centering
\caption{A sample from TAROT dataset comprising 4-tiered test cases: basic, intermediate, complex, and edge. The Reason column details the rationale for each tier assignment.}
\label{tab:sample3}
\begin{tabularx}{\textwidth}{@{} m{1.5cm} Y Y Y Y @{}}
\toprule
\multicolumn{5}{>{\hsize=\dimexpr5\hsize+8\tabcolsep\relax}Y}{
\begin{minipage}[c]{\hsize} 
\begin{alltt}
Solve the following coding problem using the programming language python:


A permutation $ - $ is a sequence of length n integers from 1 to n, in which all the numbers occur exactly once. For example, [1], [3, 5, 2, 1, 4], [1, 3, 2] $ - $ permutations, and [2, 3, 2], [4, 3, 1], [0] $ - $ no. \\

Polycarp was recently gifted a permutation a[1 \dots n] of length n. Polycarp likes trees more than permutations, so he wants to transform permutation a into a rooted binary tree. He transforms an array of different integers into a tree as follows: \\

\setlength{\leftmargini}{20pt} 
\begin{itemize}
    \item The maximum element of the array becomes the root of the tree;
    \item All elements to the left of the maximum $ - $ form a left subtree (which is built according to the same rules but applied to the left part of the array), but if there are no elements to the left of the maximum, then the root has no left child;
    \item All elements to the right of the maximum $ - $ form a right subtree (which is built according to the same rules but applied to the right side of the array), but if there are no elements to the right of the maximum, then the root has no right child.
\end{itemize} 

For example, if he builds a tree by permutation a = [3, 5, 2, 1, 4], then the root will be the element $a_2$ = 5, and the left subtree will be the tree that will be built for the subarray a[1 \dots 1] $=$ [3], and the right one $ - $ for the subarray a[3 \dots 5] $=$ [2, 1, 4]. As a result, the following tree will be built: \\

<image> The tree corresponding to the permutation a=[3, 5, 2, 1, 4]. \\

Another example: let the permutation be a=[1, 3, 2, 7, 5, 6, 4]. In this case, the tree looks like this: \\

<image> The tree corresponding to the permutation a=[1, 3, 2, 7, 5, 6, 4]. \\

Let us denote by $d_v$ the depth of the vertex $a_v$, that is, the number of edges on the path from the root to the vertex numbered $a_v$. Note that the root depth is zero. Given the permutation a, for each vertex, find the value of $d_v$. \\

Input \\

The first line contains one integer t $(1 \leq t \leq 100) \quad - $ the number of test cases. Then t test cases follow. The first line of each test case contains an integer n $(1 \leq n \leq 100) \quad - $ the length of the permutation. This is followed by n numbers $a_1, a_2, \dots, a_n \quad - $ permutation a. \\

Output \\

For each test case, output n values $ - \quad d_1, d_2, \dots, d_n$. \\

Example \\

Input \\
3 \\
5 \\
3 5 2 1 4 \\
1 \\
1 \\
4 \\
4 3 1 2 \\

Output \\
1 0 2 3 1 \\ 
0 \\
0 1 3 2 \\


The input will be stdin and you should print your solution to stdout \\

Now solve the problem and return the code. \\

\end{alltt}
\end{minipage} %
} \\
\midrule
\textbf{\small Test cases} & \textbf{\small Basic} & \textbf{\small Intermediate} & \textbf{\small Complex} & \textbf{\small Edge} \\
\midrule
\textbf{\small Input} & 
\texttt{1} \par \texttt{3} \par \texttt{1 2 3} & 
\texttt{2} \par \texttt{4} \par \texttt{2 1 4 3} \par \texttt{5} \par \texttt{5 4 3 2 1} & 
\texttt{1} \par \texttt{10} \par \texttt{3 8 2 5 10 9 1 7 4 6} & 
\texttt{1} \par \texttt{15} \par \texttt{1 2 3 4 5 6 7 8 9 10 11 12 13 14 15} \\
\addlinespace 
\textbf{\small Output} & 
\texttt{2 1 0} & 
\texttt{1 2 0 1} \par \texttt{0 1 2 3 4} & 
\texttt{2 1 3 2 0 1 3 2 4 3} & 
\texttt{14 13 12 11 10 9 8 7 6 5 4 3 2 1 0} \\
\addlinespace 
\textbf{\small Reason} & 
Simple ascending permutation forming a left‑skewed tree under normal conditions. & 
Includes a mixed permutation and a strictly decreasing permutation to test right‑skewed tree and boundary values. & 
Complex permutation of length 10 to test multiple recursion levels and both left and right subtrees. & 
Maximum ascending chain of length 15 to test deep recursion and large boundary condition. \\
\bottomrule
\end{tabularx}
\end{table*}

\clearpage
\begin{table*}[t!]
\vspace*{-0.9cm}
\fontsize{6.5}{6.5}\selectfont
\centering
\caption{A sample from TAROT dataset comprising 4-tiered test cases: basic, intermediate, complex, and edge. The Reason column details the rationale for each tier assignment.}
\label{tab:sample4}
\begin{tabularx}{\textwidth}{@{} m{1.5cm} Y Y Y Y @{}}
\toprule
\multicolumn{5}{>{\hsize=\dimexpr5\hsize+8\tabcolsep\relax}Y}{
\begin{minipage}[c]{\hsize} 
\begin{alltt}
Solve the following coding problem using the programming language python: \\

Princess'Marriage \\

Marriage of a princess \\

English text is not available in this practice contest. \\

A brave princess in a poor country, knowing that gambling payouts are determined by the parimutuel method, felt more familiar with gambling and was convinced of her victory in gambling. As a result, he spent more money than ever before and lost enough to lose all the taxes paid by the people. The King, who took this situation seriously, decided to marry the princess to the neighboring country. By doing this, I thought that I would like the princess to reflect on her daily activities and at the same time deepen her friendship with neighboring countries and receive financial assistance. \\

The princess and the prince of the neighboring country liked each other, and the kings of both countries agreed on a political marriage. The princess triumphantly went to the neighboring country with a little money in her hand. On the other hand, the motive for the princess to marry is to pursue the unilateral interests of the king, and the aide of the prince of the neighboring country who thinks that it is not pleasant shoots countless thugs along the way to make the princess dead. It was. \\

The path the princess will take has already been decided. There are a total of L post stations on the path of the princess. For convenience, the departure and arrival points are also set as post stations, and each post station is called S1, S2, \dots SL. The princess shall be in S1 first, visit the post station in ascending order (S2, S3 \dots in that order), and finally go to SL. At the post station, you can pay money to hire an escort, and as long as you have the money, you can contract for as long as you like to protect the princess. The cost of hiring an escort is 1 gold per distance. Note that the princess can also partially protect the section through which she passes. The distance between Si and Si + 1 is given by Di, and the expected value of the number of times a thug is attacked per distance between Si and Si + 1 is given by Pi. \\

Find the expected number of thugs to reach your destination when the princess has a budget of M and hires an escort to minimize the expected number of thugs. \\

Input \\
The input consists of multiple datasets. Each dataset has the following format. \\

> N M \\
> D1 P1 \\
> D2 P2 \\
> ... \\
> DN PN \\

Two integers are given in the first row of each dataset, representing the number of intervals N $(1 \leq N \leq 10,000$ and the budget M $(0 \leq M \leq 1,000,000,000$ of the princess difference, respectively. The next N lines show information about the path the princess takes. Each line contains two integers, and the i-th line is the expected value of the interval Di $(1 \leq Di \leq 10,000)$ and the number of attacks when moving one unit distance between them Pi $(0 \leq Pi \leq 10)$ ). The end of the input is represented by a data set with N = 0 and M = 0. Do not output the calculation result for this data set. \\

Output

For each dataset, output the expected number of times the princess will be attacked by thugs to your destination. \\

Sample Input \\
2 8 \\
5 6 \\
4 5 \\
3 1 \\
5 10 \\
5 10 \\
5 10 \\
0 0 \\

Output for the Sample Input \\
Five \\
140 \\

The input will be given via stdin and the output should be printed to stdout by your code. 
\end{alltt}

\end{minipage} %
} \\
\midrule
\textbf{\small Test cases} & \textbf{\small Basic} & \textbf{\small Intermediate} & \textbf{\small Complex} & \textbf{\small Edge} \\
\midrule
\textbf{\small Input} & 
\texttt{4 7} \par \texttt{3 2} \par \texttt{4 1} \par \texttt{1 5} \par \texttt{2 2} \par \texttt{0 0} & 
\texttt{6 12} \par \texttt{5 3} \par \texttt{2 0} \par \texttt{7 2} \par \texttt{3 3} \par \texttt{4 1} \par \texttt{6 2} \par \texttt{0 0} & 
\texttt{10 30} \par \texttt{10 1} \par \texttt{5 5} \par \texttt{8 5} \par \texttt{6 3} \par \texttt{12 2} \par \texttt{4 5} \par \texttt{7 3} \par \texttt{9 4} \par \texttt{3 0} \par \texttt{11 2} \par \texttt{0 0} & 
\texttt{3 0} \par \texttt{5 2} \par \texttt{10 4} \par \texttt{7 3} \par \texttt{4 100} \par \texttt{5 2} \par \texttt{10 4} \par \texttt{7 3} \par \texttt{8 0} \par \texttt{2 1000} \par \texttt{100 0} \par \texttt{200 0} \par \texttt{0 0} \\
\addlinespace 
\textbf{\small Output} & 
\texttt{3} & 
\texttt{22} & 
\texttt{83} & 
\texttt{71} \par \texttt{0} \par \texttt{0} \\
\addlinespace 
\textbf{\small Reason} & 
Simple scenario with multiple segments and straightforward positive Pi values; tests basic greedy coverage under a limited budget. & 
Moderate number of segments including Pi=0, ensuring segments with no attacks are ignored and budget partially covers higher-Pi segments. & 
Larger set of segments with ties in Pi values and varied distances, requiring correct sorting and partial coverage among equal-Pi segments. & 
Edge conditions including zero budget, budget exceeding total distance, and segments with Pi=0 to verify no-protection and full-protection behaviors. \\
\bottomrule
\end{tabularx}
\end{table*}

\clearpage
\begin{table*}[t!]
\vspace*{-0.4cm}
\fontsize{7}{8.5}\selectfont
\centering
\caption{A sample from TAROT dataset comprising 4-tiered test cases: basic, intermediate, complex, and edge. The Reason column details the rationale for each tier assignment.}
\label{tab:sample5}
\begin{tabularx}{\textwidth}{@{} m{1.5cm} Y Y Y Y @{}}
\toprule
\multicolumn{5}{>{\hsize=\dimexpr5\hsize+8\tabcolsep\relax}Y}{
\begin{minipage}[c]{\hsize} 
\begin{alltt}
Solve the following coding problem using the programming language python: \\

Valera loves his garden, where n fruit trees grow. \\

This year he will enjoy a great harvest! On the i-th tree $b_i$ fruit grow, they will ripen on a day number $a_i$. Unfortunately, the fruit on the tree get withered, so they can only be collected on day $a_i$ and day $a_i + 1$ (all fruits that are not collected in these two days, become unfit to eat). \\

Valera is not very fast, but there are some positive points. Valera is ready to work every day. In one day, Valera can collect no more than v fruits. The fruits may be either from the same tree, or from different ones. What is the maximum amount of fruit Valera can collect for all time, if he operates optimally well? \\

-----Input----- \\
The first line contains two space-separated integers n and v $(1 \leq n, v \leq 3000)$ $ - $ the number of fruit trees in the garden and the number of fruits that Valera can collect in a day. \\

Next n lines contain the description of trees in the garden. The i-th line contains two space-separated integers $a_i$ and $b_i$ $(1 \leq a_i, b_i \leq 3000)$ $ - $ the day the fruits ripen on the i-th tree and the number of fruits on the i-th tree. \\

-----Output----- \\
Print a single integer $ - $ the maximum number of fruit that Valera can collect.  \\

-----Examples----- \\
Input \\
2 3 \\
1 5 \\
2 3 \\

Output \\
8 \\

Input \\
5 10 \\
3 20 \\
2 20 \\
1 20 \\
4 20 \\
5 20 \\

Output \\
60 \\

-----Note----- \\
In the first sample, in order to obtain the optimal answer, you should act as follows.   On the first day collect 3 fruits from the 1-st tree.  On the second day collect 1 fruit from the 2-nd tree and 2 fruits from the 1-st tree.  On the third day collect the remaining fruits from the 2-nd tree. \\

In the second sample, you can only collect 60 fruits, the remaining fruit will simply wither. \\

The input will be stdin and you should print your solution to stdout \\

Now solve the problem and return the code.
\end{alltt}

\end{minipage} %
} \\
\midrule
\textbf{\small Test cases} & \textbf{\small Basic} & \textbf{\small Intermediate} & \textbf{\small Complex} & \textbf{\small Edge} \\
\midrule
\textbf{\small Input} & 
\texttt{2 5} \par \texttt{1 3} \par \texttt{3 4} & 
\texttt{3 1} \par \texttt{1 2} \par \texttt{2 2} \par \texttt{3 2}& 
\texttt{5 5} \par \texttt{1 4} \par \texttt{2 6} \par \texttt{2 3} \par \texttt{5 10} \par \texttt{6 2} & 
\texttt{2 1000} \par \texttt{2999 1500} \par \texttt{3000 2500} \\
\addlinespace 
\textbf{\small Output} & 
\texttt{7} & 
\texttt{4} & 
\texttt{25} & 
\texttt{3000} \\
\addlinespace 
\textbf{\small Reason} & 
No overlapping ripening days and capacity exceeds daily fruits; collect all fruits on their ripening days. & 
Capacity is only 1 per day with overlapping two-day windows; requires optimal scheduling across consecutive days. & 
Multiple trees ripen on the same days, gaps between ripening days, and moderate capacity to stress multi-day planning. & 
Ripening on the maximum allowed days (2999 and 3000) tests boundary handling and two-day collection windows at the end of the range. \\
\bottomrule
\end{tabularx}
\end{table*}

\clearpage
\begin{table*}[t!]
\vspace*{-0.4cm}
\fontsize{7.5}{8.5}\selectfont
\centering
\caption{A sample from TAROT dataset comprising 4-tiered test cases: basic, intermediate, complex, and edge. The Reason column details the rationale for each tier assignment.}
\label{tab:sample6}
\begin{tabularx}{\textwidth}{@{} m{1.5cm} Y Y Y Y @{}}
\toprule
\multicolumn{5}{>{\hsize=\dimexpr5\hsize+8\tabcolsep\relax}Y}{
\begin{minipage}[c]{\hsize} 
\begin{alltt}
Solve the following coding problem using the programming language python: \\

Bessie the cow has just intercepted a text that Farmer John sent to Burger Queen! However, Bessie is sure that there is a secret message hidden inside. \\

The text is a string $s$ of lowercase Latin letters. She considers a string $t$ as hidden in string $s$ if $t$ exists as a subsequence of $s$ whose indices form an arithmetic progression. For example, the string aab is hidden in string aaabb because it occurs at indices $1$, $3$, and $5$, which form an arithmetic progression with a common difference of $2$. Bessie thinks that any hidden string that occurs the most times is the secret message. Two occurrences of a subsequence of $S$ are distinct if the sets of indices are different. Help her find the number of occurrences of the secret message! \\

For example, in the string aaabb, a is hidden $3$ times, b is hidden $2$ times, ab is hidden $6$ times, aa is hidden $3$ times, bb is hidden $1$ time, aab is hidden $2$ times, aaa is hidden $1$ time, abb is hidden $1$ time, aaab is hidden $1$ time, aabb is hidden $1$ time, and aaabb is hidden $1$ time. The number of occurrences of the secret message is $6$. \\

-----Input----- \\

The first line contains a string $s$ of lowercase Latin letters ($1 \leq |s| \leq 10^{5}$) — the text that Bessie intercepted. \\

-----Output----- \\

Output a single integer $ - $ the number of occurrences of the secret message. \\

-----Examples----- \\
Input \\
aaabb \\

Output \\
6 \\

Input \\
usaco \\

Output \\
1 \\

Input \\
lol \\

Output \\
2 \\

-----Note----- \\

In the first example, these are all the hidden strings and their indice sets:   a occurs at $(1)$, $(2)$, $(3)$  b occurs at $(4)$, $(5)$  ab occurs at $(1,4)$, $(1,5)$, $(2,4)$, $(2,5)$, $(3,4)$, $(3,5)$  aa occurs at $(1,2)$, $(1,3)$, $(2,3)$  bb occurs at $(4,5)$  aab occurs at $(1,3,5)$, $(2,3,4)$  aaa occurs at $(1,2,3)$  abb occurs at $(3,4,5)$  aaab occurs at $(1,2,3,4)$  aabb occurs at $(2,3,4,5)$  aaabb occurs at $(1,2,3,4,5)$  Note that all the sets of indices are arithmetic progressions. \\

In the second example, no hidden string occurs more than once. \\

In the third example, the hidden string is the letter l. \\

The input will be stdin and you should print your solution to stdout \\

Now solve the problem and return the code. \\
\end{alltt}

\end{minipage} %
} \\
\midrule
\textbf{\small Test cases} & \textbf{\small Basic} & \textbf{\small Intermediate} & \textbf{\small Complex} & \textbf{\small Edge} \\
\midrule
\textbf{\small Input} & 
\texttt{abab} & 
\texttt{abacaba} & 
\texttt{abababab} & 
\texttt{z} \\
\addlinespace 
\textbf{\small Output} & 
\texttt{3} & 
\texttt{6} & 
\texttt{10} & 
\texttt{1} \\
\addlinespace 
\textbf{\small Reason} & 
A simple alternating pattern to validate basic subsequence counting. & 
Mixed letters and repeating patterns to test moderately complex subsequences. & 
Longer alternating pattern to stress test counting of many arithmetic‐progression subsequences. & 
Minimal input length boundary case. \\
\bottomrule
\end{tabularx}
\end{table*}

\clearpage
\begin{table*}[t!]
\vspace*{-0.9cm}
\fontsize{6}{6.05}\selectfont
\centering
\caption{A sample from TAROT dataset comprising 4-tiered test cases: basic, intermediate, complex, and edge. The Reason column details the rationale for each tier assignment.}
\label{tab:sample7}
\begin{tabularx}{\textwidth}{@{} m{1.5cm} Y Y Y Y @{}}
\toprule
\multicolumn{5}{>{\hsize=\dimexpr5\hsize+8\tabcolsep\relax}Y}{
\begin{minipage}[c]{\hsize} 
\begin{alltt}
Solve the following coding problem using the programming language python: \\

The ZCO scholarship contest offers scholarships to first time ZCO participants. You are participating in it for the first time. So you want to know the number of participants who'll get the scholarship. You know that the maximum number of scholarships offered is $R$ and there are a total of $N$ participants numbered from $1$ to $N$. Out of these, you know the set of people (denoted by $X$) who you know, had participated in previous year ZCOs and hence, they shall not get the scholarship. Further, as the world isn't free from plagiarism, so is the case with the scholarship contest. And from your secret sources, you also know the set of people (denoted by set $Y$) who were involved in plagiarism and therefore aren't eligible for scholarship either. \\

Find out the number of participants who shall get the scholarship. \\

PS: Don't ask how so many scholarships are being offered when you see the constraints on $R$. You never questioned it when in mathematics classes, some person bought $80$ watermelons twice just to compare them and save 1. \\ %\rupee~

-----Input:----- \\
- The first line will contain a single integer, $T$, the number of testcases. Then the testcases follow.  \\
- The first line of each test case contains four integers; $N$, $R$, $|X|$ and $|Y|$ denoting the number of participants, maximum number of scholarships offered, number of old participants, and the number of participants involved in plagiarism, respectively. \\
- The second line of each test case contains $|X|$ space separated integers $x_1, x_2 \ldots x_{|X|}$ denoting the indices of people who participated in previous years. If $X$ is empty, this line is skipped and no empty line is in the input. \\
- The third line of each test case contains $|Y|$ space separated integers $y_1, y_2 \ldots y_{|Y|}$  denoting the indices of people who are involved in plagiarism. If $Y$ is empty, this line is skipped and no empty line is in input. \\

-----Output:----- \\
For each testcase, print a single integer in a new line, denoting the number of participants who shall get the scholarship. \\

-----Constraints----- \\
- $1 \leq T \leq 1000$ \\
- $1 \leq N \leq 10^{15}$ \\
- $0 \leq R \leq 10^{15}$ \\
- $0 \leq |X|, |Y| \leq min(N, 2*10^5)$ \\
- $1 \leq x_i, y_i \leq N$ \\
- All $x_i$ are distinct \\
- All $y_i$ are distinct \\
- Sum of $|X|$ over all test cases does not exceed $5*10^5$ \\
- Sum of $|Y|$ over all test cases does not exceed $5*10^5$ \\

-----Subtasks----- \\
- 20 points : $1 \leq N \leq 10^3$, and the sum of $N$ over all test cases does not exceed $3*10^3$ \\
- 30 points : $1 \leq N \leq 2*10^5$, and the sum of $N$ over all test cases does not exceed $5*10^5$ \\
- 50 points: Original constraints \\

-----Sample Input:----- \\
3 \\
5 3 0 1 \\
4 \\
10 2 4 6 \\
3 1 7 6 \\
4 3 1 5 9 7 \\
10 4 4 6 \\
3 1 7 6 \\
4 3 1 5 9 7 \\

-----Sample Output:----- \\
3 \\
2 \\
3 \\

-----EXPLANATION:----- \\
- In the first testcase, only participant $4$ is involved in plagiarism, and thus not eligible for the scholarship. No user has participated in previous years, and so no empty line is there in the sample. All participants except participant $4$ are eligible for the scholarship, but only three of them get it because $R = 3$. \\
- Both second and third testcases are the same, except for $R$. In both samples, only participants $2$, $8$ and $10$ are eligible for scholarships. \\
- In the second testcase, since the maximum number of scholarships is $2$, only $2$ participants get scholarships. \\
- In the third testcase, all three eligible participants get scholarships. \\

The input will be stdin and you should print your solution to stdout
\end{alltt}

\end{minipage} %
} \\
\midrule
\textbf{\small Test cases} & \textbf{\small Basic} & \textbf{\small Intermediate} & \textbf{\small Complex} & \textbf{\small Edge} \\
\midrule
\textbf{\small Input} & 
\texttt{1} \par \texttt{4 2 1 1} \par \texttt{1} \par \texttt{4} & 
\texttt{2} \par \texttt{7 4 3 2} \par \texttt{2 4 5} \par \texttt{4 6} \par \texttt{5 1 0 2} \par \texttt{3 5} &
\texttt{3} \par \texttt{1000000000000 1000000000000 2 3} \par \texttt{1 1000000000000} \par \texttt{500000000000 1 999999999999} \par \texttt{20 15 5 5} \par \texttt{1 2 3 4 5} \par \texttt{4 5 6 7 8} \par \texttt{50 100 3 2} \par \texttt{10 20 30} \par \texttt{30 40} & 
\texttt{2} \par \texttt{1000000000000000 0 0 0} \par \texttt{5 10 3 3} \par \texttt{1 2 3} \par \texttt{3 4 5} \\
\addlinespace 
\textbf{\small Output} & 
\texttt{2} & 
\texttt{3} \par \texttt{1} & 
\texttt{999999999996} \par \texttt{12} \par \texttt{46} &
\texttt{0} \par \texttt{0} \\
\addlinespace 
\textbf{\small Reason} & 
Basic test with a single test case, non-empty X and Y sets without overlap, validating core functionality. & 
Medium complexity with multiple test cases, overlapping X and Y in the first, and an empty X set in the second. & 
High complexity with very large N and R values, moderate X and Y sizes, and multiple test cases to stress-test the implementation. & 
Edge case with maximum boundary values and zero scholarships in the first, and X and Y covering all participants in the second, testing empty sets and full exclusion. \\
\bottomrule
\end{tabularx}
\end{table*}

\clearpage
\begin{table*}[t!]
\vspace*{-0.7cm}
\fontsize{6}{6.5}\selectfont
\centering
\caption{A sample from TAROT dataset comprising 4-tiered test cases: basic, intermediate, complex, and edge. The Reason column details the rationale for each tier assignment.}
\label{tab:sample7}
\begin{tabularx}{\textwidth}{@{} m{1.5cm} Y Y Y Y @{}}
\toprule
\multicolumn{5}{>{\hsize=\dimexpr5\hsize+8\tabcolsep\relax}Y}{
\begin{minipage}[c]{\hsize} 
\begin{alltt}
Solve the following coding problem using the programming language python: \\

Polycarp has recently got himself a new job. He now earns so much that his old wallet can't even store all the money he has. Berland bills somehow come in lots of different sizes. However, all of them are shaped as rectangles (possibly squares). All wallets are also produced in form of rectangles (possibly squares). \\

A bill $x \times y$ fits into some wallet $h \times w$ if either $x \le h$ and $y \le w$ or $y \le h$ and $x \le w$. Bills can overlap with each other in a wallet and an infinite amount of bills can fit into a wallet. That implies that all the bills Polycarp currently have fit into a wallet if every single one of them fits into it independently of the others. \\

Now you are asked to perform the queries of two types: \\

  $+~x~y$ — Polycarp earns a bill of size $x \times y$;  $?~h~w$ — Polycarp wants to check if all the bills he has earned to this moment fit into a wallet of size $h \times w$.  \\

It is guaranteed that there is at least one query of type $1$ before the first query of type $2$ and that there is at least one query of type $2$ in the input data. For each query of type $2$ print "YES" if all the bills he has earned to this moment fit into a wallet of given size. Print "NO" otherwise. \\

-----Input----- \\

The first line contains a single integer $n$ ($2 \le n \le 5 \cdot 10^5$) — the number of queries. \\

Each of the next $n$ lines contains a query of one of these two types: \\

  $+~x~y$ ($1 \le x, y \le 10^9$) — Polycarp earns a bill of size $x \times y$;  $?~h~w$ ($1 \le h, w \le 10^9$) — Polycarp wants to check if all the bills he has earned to this moment fit into a wallet of size $h \times w$.  \\

It is guaranteed that there is at least one query of type $1$ before the first query of type $2$ and that there is at least one query of type $2$ in the input data. \\

-----Output----- \\

For each query of type $2$ print "YES" if all the bills he has earned to this moment fit into a wallet of given size. Print "NO" otherwise. \\

-----Example----- \\
Input \\
9 \\
+ 3 2 \\
+ 2 3 \\
? 1 20 \\
? 3 3 \\
? 2 3 \\
+ 1 5 \\
? 10 10 \\
? 1 5 \\
+ 1 1 \\

Output \\
NO \\
YES \\
YES \\
YES \\
NO \\

-----Note----- \\

The queries of type $2$ of the example: \\

  Neither bill fits;  Both bills fit (just checking that you got that bills can overlap);  Both bills fit (both bills are actually the same);  All bills fit (too much of free space in a wallet is not a problem);  Only bill $1 \times 5$ fit (all the others don't, thus it's "NO"). \\

The input will be stdin and you should print your solution to stdout
\end{alltt}

\end{minipage} %
} \\
\midrule
\textbf{\small Test cases} & \textbf{\small Basic} & \textbf{\small Intermediate} & \textbf{\small Complex} & \textbf{\small Edge} \\
\midrule
\textbf{\small Input} & 
\texttt{4} \par \texttt{+ 4 5} \par \texttt{? 5 4} \par \texttt{? 4 4} \par \texttt{? 6 4} & 
\texttt{7} \par \texttt{+ 2 7} \par \texttt{+ 3 3} \par \texttt{+ 7 2} \par \texttt{? 3 7} \par \texttt{? 4 6} \par \texttt{+ 10 1} \par \texttt{? 10 5} & 
\texttt{13} \par \texttt{+ 5 5} \par \texttt{+ 6 4} \par \texttt{+ 9 1} \par \texttt{? 5 5} \par \texttt{? 9 1} \par \texttt{? 4 9} \par \texttt{+ 2 8} \par \texttt{? 8 9} \par \texttt{+ 7 7} \par \texttt{? 7 7} \par \texttt{? 8 7} \par \texttt{? 10 8} \par \texttt{? 6 6} &
\texttt{8} \par \texttt{+ 1000000000 1} \par \texttt{+ 1 1000000000} \par \texttt{+ 500000000 500000000} \par \texttt{? 1000000000 1000000000} \par \texttt{? 999999999 1000000000} \par \texttt{? 1000000000 499999999} \par \texttt{+ 1000000000 1000000000} \par \texttt{? 1000000000 1000000000} \\

\addlinespace 
\textbf{\small Output} & 
\texttt{YES} \par \texttt{NO} \par \texttt{YES} & 
\texttt{YES} \par \texttt{NO} \par \texttt{YES} & 
\texttt{NO} \par \texttt{NO} \par \texttt{NO} \par \texttt{YES} \par \texttt{NO} \par \texttt{NO} \par \texttt{YES} \par \texttt{NO} & 
\texttt{YES} \par \texttt{YES} \par \texttt{NO} \par \texttt{YES} \\

\addlinespace 
\textbf{\small Reason} & 
Single bill with queries testing orientation and size validation under straightforward conditions. & 
Multiple bills including duplicates and interleaved adds and queries testing correct global dimension tracking. & 
Complex interleaving of many adds and queries with varying dimensions to stress test global maximum computations. & 
Extreme boundary values testing maximum limits and strict comparison edge where one dimension is just below requirement. \\
\bottomrule
\end{tabularx}
\end{table*}

\clearpage
\begin{table*}[t!]
\vspace*{-0.7cm}
\fontsize{7}{7.5}\selectfont
\centering
\caption{A sample from TAROT dataset comprising 4-tiered test cases: basic, intermediate, complex, and edge. The Reason column details the rationale for each tier assignment.}
\label{tab:sample7}
\begin{tabularx}{\textwidth}{@{} m{1.5cm} Y Y Y Y @{}}
\toprule
\multicolumn{5}{>{\hsize=\dimexpr5\hsize+8\tabcolsep\relax}Y}{
\begin{minipage}[c]{\hsize} 
\begin{alltt}
Solve the following coding problem using the programming language python: \\

Valera had an undirected connected graph without self-loops and multiple edges consisting of n vertices. The graph had an interesting property: there were at most k edges adjacent to each of its vertices. For convenience, we will assume that the graph vertices were indexed by integers from 1 to n. \\

One day Valera counted the shortest distances from one of the graph vertices to all other ones and wrote them out in array d. \\

Thus, element d[i] of the array shows the shortest distance from the vertex Valera chose to vertex number i. \\

Then something irreparable terrible happened. Valera lost the initial graph. However, he still has the array d. Help him restore the lost graph. \\

Input \\

The first line contains two space-separated integers n and k $(1 \leq k \leq 105)$. Number n shows the number of vertices in the original graph. Number k shows that at most k edges were adjacent to each vertex in the original graph. \\

The second line contains space-separated integers d[1], d[2], ..., d[n] $(0 \leq d[i] < n)$. Number d[i] shows the shortest distance from the vertex Valera chose to the vertex number i. \\

Output \\

If Valera made a mistake in his notes and the required graph doesn't exist, print in the first line number -1. Otherwise, in the first line print integer m $(0 \leq m \leq 106)$ $ - $ the number of edges in the found graph. \\

In each of the next m lines print two space-separated integers ai and bi $(1 \leq ai, \quad bi \leq n; \quad ai \neq bi)$, denoting the edge that connects vertices with numbers ai and bi. The graph shouldn't contain self-loops and multiple edges. If there are multiple possible answers, print any of them. \\

Examples \\

Input \\
3 2 \\
0 1 1 \\

Output \\
3 \\
1 2 \\
1 3 \\
3 2 \\

Input \\
4 2 \\
2 0 1 3 \\

Output \\
3 \\
1 3 \\
1 4 \\
2 3 \\

Input \\
3 1 \\
0 0 0 \\

Output \\
-1 \\

The input will be given via stdin and the output should be printed to stdout by your code.
\end{alltt}

\end{minipage} %
} \\
\midrule
\textbf{\small Test cases} & \textbf{\small Basic} & \textbf{\small Intermediate} & \textbf{\small Complex} & \textbf{\small Edge} \\
\midrule
\textbf{\small Input} & 
\texttt{4 2} \par \texttt{0 1 1 2} &
\texttt{7 3} \par \texttt{0 1 2 2 1 2 3} & 
\texttt{10 3} \par \texttt{0 1 1 1 2 2 2 2 2 3} &
\texttt{5 3} \par \texttt{0 2 2 3 3} \\

\addlinespace 
\textbf{\small Output} & 
\texttt{3} \par \texttt{1 2} \par \texttt{1 3} \par \texttt{2 4} & 
\texttt{6} \par \texttt{1 2} \par \texttt{1 5} \par \texttt{2 3} \par \texttt{2 4} \par \texttt{5 6} \par \texttt{3 7} & 
\texttt{9} \par \texttt{1 2} \par \texttt{1 3} \par \texttt{1 4} \par \texttt{2 5} \par \texttt{2 6} \par \texttt{3 7} \par \texttt{3 8} \par \texttt{4 9} \par \texttt{5 10} & 
\texttt{-1} \\

\addlinespace 
\textbf{\small Reason} & 
Simple BFS tree with one level-2 vertex. & 
Moderately sized tree with branching and various depths. & 
Larger tree with multiple branches and depth-3 leaf. & 
No vertices at distance 1, invalid distance sequence \\
\bottomrule
\end{tabularx}
\end{table*}

\clearpage
\begin{table*}[t!]
\vspace*{-0.5cm}
\fontsize{7.5}{7.5}\selectfont
\centering
\caption{A sample from TAROT dataset comprising 4-tiered test cases: basic, intermediate, complex, and edge. The Reason column details the rationale for each tier assignment.}
\label{tab:sample7}
\begin{tabularx}{\textwidth}{@{} m{1.5cm} Y Y Y Y @{}}
\toprule
\multicolumn{5}{>{\hsize=\dimexpr5\hsize+8\tabcolsep\relax}Y}{
\begin{minipage}[c]{\hsize} 
\begin{alltt}
Solve the following coding problem using the programming language python: \\

The game of Berland poker is played with a deck of $n$ cards, $m$ of which are jokers. $k$ players play this game ($n$ is divisible by $k$). \\

At the beginning of the game, each player takes $\frac{n}{k}$ cards from the deck (so each card is taken by exactly one player). The player who has the maximum number of jokers is the winner, and he gets the number of points equal to $x - y$, where $x$ is the number of jokers in the winner's hand, and $y$ is the maximum number of jokers among all other players. If there are two or more players with maximum number of jokers, all of them are winners and they get $0$ points. \\

Here are some examples:  $n = 8$, $m = 3$, $k = 2$. If one player gets $3$ jokers and $1$ plain card, and another player gets $0$ jokers and $4$ plain cards, then the first player is the winner and gets $3 - 0 = 3$ points;  $n = 4$, $m = 2$, $k = 4$. Two players get plain cards, and the other two players get jokers, so both of them are winners and get $0$ points;  $n = 9$, $m = 6$, $k = 3$. If the first player gets $3$ jokers, the second player gets $1$ joker and $2$ plain cards, and the third player gets $2$ jokers and $1$ plain card, then the first player is the winner, and he gets $3 - 2 = 1$ point;  $n = 42$, $m = 0$, $k = 7$. Since there are no jokers, everyone gets $0$ jokers, everyone is a winner, and everyone gets $0$ points.  \\

Given $n$, $m$ and $k$, calculate the maximum number of points a player can get for winning the game. \\

-----Input----- \\

The first line of the input contains one integer $t$ ($1 \le t \le 500$) — the number of test cases. \\

Then the test cases follow. Each test case contains three integers $n$, $m$ and $k$ ($2 \le n \le 50$, $0 \le m \le n$, $2 \le k \le n$, $k$ is a divisors of $n$). \\

-----Output----- \\

For each test case, print one integer — the maximum number of points a player can get for winning the game. \\

-----Example----- \\
Input \\
4 \\
8 3 2 \\
4 2 4 \\
9 6 3 \\
42 0 7 \\

Output \\
3 \\
0 \\
1 \\
0 \\

-----Note----- \\

Test cases of the example are described in the statement. \\

The input will be stdin and you should print your solution to stdout \\

Now solve the problem and return the code. \\
\end{alltt}

\end{minipage} %
} \\
\midrule
\textbf{\small Test cases} & \textbf{\small Basic} & \textbf{\small Intermediate} & \textbf{\small Complex} & \textbf{\small Edge} \\
\midrule
\textbf{\small Input} & 
\texttt{3} \par \texttt{9 2 3} \par \texttt{12 5 4} \par \texttt{6 5 3} &
\texttt{5} \par \texttt{20 10 5} \par \texttt{15 3 5} \par \texttt{10 10 2} \par \texttt{14 7 7} \par \texttt{18 17 3} & 
\texttt{7} \par \texttt{50 25 25} \par \texttt{49 49 7} \par \texttt{48 20 6} \par \texttt{30 0 5} \par \texttt{32 16 4} \par \texttt{45 23 9} \par \texttt{28 14 7} &
\texttt{6} \par \texttt{2 0 2} \par \texttt{2 2 2} \par \texttt{50 0 25} \par \texttt{50 50 50} \par \texttt{50 25 5} \par \texttt{50 1 2} \\

\addlinespace 
\textbf{\small Output} & 
\texttt{2} \par \texttt{2} \par \texttt{0} & 
\texttt{2} \par \texttt{3} \par \texttt{0} \par \texttt{1} \par \texttt{0} & 
\texttt{1} \par \texttt{0} \par \texttt{5} \par \texttt{0} \par \texttt{5} \par \texttt{2} \par \texttt{2} & 
\texttt{0} \par \texttt{0} \par \texttt{0} \par \texttt{0} \par \texttt{6} \par \texttt{1} \\

\addlinespace 
\textbf{\small Reason} & 
Simple small cases covering scenarios where jokers are fewer than, equal to, or exceed the per-player limit. & 
Moderate-sized inputs, testing exact division of jokers, no jokers, and tied maximum distributions. & 
Varied larger values including big decks, testing heavy distributions and zero-joker scenarios. & 
Extreme boundary conditions with minimal and maximal n, k, and m values to test edge handling. \\
\bottomrule
\end{tabularx}
\end{table*}





\end{document}